%% file: main.tex
\documentclass{article}

\PassOptionsToPackage{table,HTML}{xcolor}
\usepackage{CJKutf8}
\usepackage{meituan_style}

\usepackage{adjustbox}
\usepackage{algorithm}
\usepackage[utf8]{inputenc} %
\usepackage[T1]{fontenc}    %
\usepackage{newunicodechar} %
\usepackage[normalem]{ulem} %
\usepackage{float}
\usepackage{url}
\usepackage{setspace}
\usepackage{pifont,fontawesome5}
\usepackage{booktabs}
\usepackage{amsfonts}
\usepackage{nicefrac}
\usepackage{microtype}
\usepackage{lipsum}
\usepackage{natbib}
\usepackage{doi}
\usepackage{placeins}
\usepackage{enumitem}
\usepackage{algorithmic}
\usepackage{array}
\usepackage{minitoc}
\usepackage{tabularx}
\usepackage{listings}
\usepackage{cancel}
\usepackage{multirow}
\usepackage{multicol}
\usepackage{makecell}

\usepackage{caption}
\usepackage{subcaption}
\usepackage{graphicx}
\usepackage[inkscapelatex=false]{svg}
\usepackage{tikz}
\graphicspath{{figures/}}
\usepackage{wrapfig}
\definecolor{groupgray}{RGB}{230,232,235}
\definecolor{textgray}{RGB}{140,140,140}

\usepackage{amssymb}
\usepackage{amsmath} 

\usepackage{hyperref, cleveref}
\hypersetup{
    colorlinks=true,      %
    linkcolor=meituanyellow,      %
    urlcolor=meituanyellow,       %
    citecolor=meituanyellow,      %
    linkbordercolor=meituanyellow, %
    urlbordercolor=meituanyellow,
    citebordercolor=meituanyellow,
    pdfborderstyle={/S/U/W 1}, %
}

\setlist[itemize]{leftmargin=*}
\setlist[enumerate]{leftmargin=*}
\setlist[description]{leftmargin=*}

\newcommand{\MRAlgFunc}[1]{\text{\textsc{#1}}}

\title{Meituan-Robotics-0: A Vision-Language-Action Foundation Model for Desktop Manipulation}

\author{Meituan Robotics Team\thanks{See \hyperref[sec:contributions]{Contributions} for the full author list.}}

\renewcommand{\shorttitle}{Meituan-Robotics-0}

\begin{document}
\maketitle
\input{sections/00-abstract}

\input{sections/01-introduction}
\input{sections/02-data_curation}

\input{sections/03-model}

\input{sections/04-deployment}
\input{sections/06-experiments}
\input{sections/07-conclusion}

\clearpage
\input{sections/08-contributions}

\clearpage

\input{sections/appendix}

\bibliographystyle{unsrt}
\bibliography{references}

\end{document}

%% file: sections/00-abstract.tex
\begin{abstract}

General-purpose desktop manipulation requires an embodied system that can acquire broad behavioral priors from heterogeneous data, transfer them across robot embodiments, recover from closed-loop execution errors, and preserve task-relevant state over extended horizons. We present Meituan-Robotics-0 (MR0), a vision-language-action foundation model and embodied-agent system for stationary-base desktop manipulation. Its data foundation combines 147 million vision-language samples with 15,060 hours of curated robot trajectories spanning 12 embodiments.
MR0 converts this heterogeneous experience into executable control through a three-stage training recipe. Discrete action-token pre-training first aligns multimodal representations with robot behavior. Continuous flow-matching mid-training then transfers these priors to fine-grained cross-embodiment control. Real-world post-training further adapts the policy using successful demonstrations, human-gated corrections, and outcome-labeled autonomous rollouts. 
We introduce Rollback, a data collection mechanism that pairs replaced policy segments with human corrections for CFG-DAgger training. A unified end-effector interface supports learning across heterogeneous embodiments, while asynchronous action-chunk merging maintains smooth execution under variable inference latency. Because the low-level policy is reactive, an event-driven agent converts visual observations into persistent task events and invokes task-level reassessment only when task progress or feasibility changes. This separation preserves continuous motor control while enabling cross-task retrieval and conditional task resumption.
Evaluated as a standalone policy, MR0 achieves 93.6\% success rate on RoboTwin~2.0 and 15.73\% success rate on RoboDojo. Under matched HG-DAgger post-training, MR0 achieves higher average success rate and progress score than $\pi_{0.5}$ across the three tasks.
On Ethernet insertion, CFG-DAgger trained with Rollback data reaches 70\% success and a progress score of 0.90 without value learning or advantage labeling.
Qualitative agent deployments further demonstrate cross-task event retrieval and recovery from temporarily unsatisfied task conditions. These results show that aligned heterogeneous data, real-world post-training, and event-level memory provide complementary foundations for transferable and long-horizon desktop manipulation.

\par\vspace{0.5em}\noindent\makebox[\linewidth][c]{\small\normalfont \textcolor[HTML]{444444}{\faHome}~\href{https://meituan-robotics.github.io/Meituan-Robotics-0/}{\textcolor[HTML]{B38600}{\texttt{Homepage}}}\quad\textcolor[HTML]{444444}{\textbar}\quad \textcolor[HTML]{444444}{\faGithub}~\href{https://github.com/Meituan-Robotics/Meituan-Robotics-0}{\textcolor[HTML]{B38600}{\texttt{GitHub}}}\quad\textcolor[HTML]{444444}{\textbar}\quad \textcolor[HTML]{444444}{\faRobot}~\href{https://huggingface.co/MeituanRobotics/models}{\textcolor[HTML]{B38600}{\texttt{Model}}}\quad\textcolor[HTML]{444444}{\textbar}\quad \textcolor[HTML]{444444}{\faDatabase}~\href{https://huggingface.co/datasets/MeituanRobotics/RoboDesk}{\textcolor[HTML]{B38600}{\texttt{Data}}}}\par\end{abstract}

%% file: sections/01-introduction.tex
\section{Introduction}

Vision--language--action~\cite{internvla_a1_2026, ding2026skillplug,internvla_a15_2607,qwenvla2026,starvla2026,lingbotvla2026,xvla2026} models provide a general interface for mapping visual observations and language instructions to robot actions. Their increasing data and model scales have substantially broadened the range of manipulation behaviors that can be learned within a single policy. Deploying such policies on desktop manipulation, however, exposes several limitations that cannot be addressed by scaling the action model alone. Robot datasets~\cite{agibotworld,galaxea,khazatsky2024droid,tian2025interndata,robocoin2025,zhang2026hy,wang2026robointer1} collected from different sources use inconsistent observation formats, action spaces, embodiments, and task descriptions, which prevents them from providing coherent supervision without careful alignment. Even after data alignment, policies trained primarily on successful demonstrations receive little guidance for recovering from the error states induced by their own closed-loop behavior. Moreover, a reactive policy conditioned only on the current observation cannot preserve task information once it leaves the observation window. A practical desktop-manipulation system must therefore connect aligned heterogeneous data, robust real-world policy learning, and persistent task-level memory.

\begin{figure}[t]
    \centering
\includegraphics[width=\linewidth]{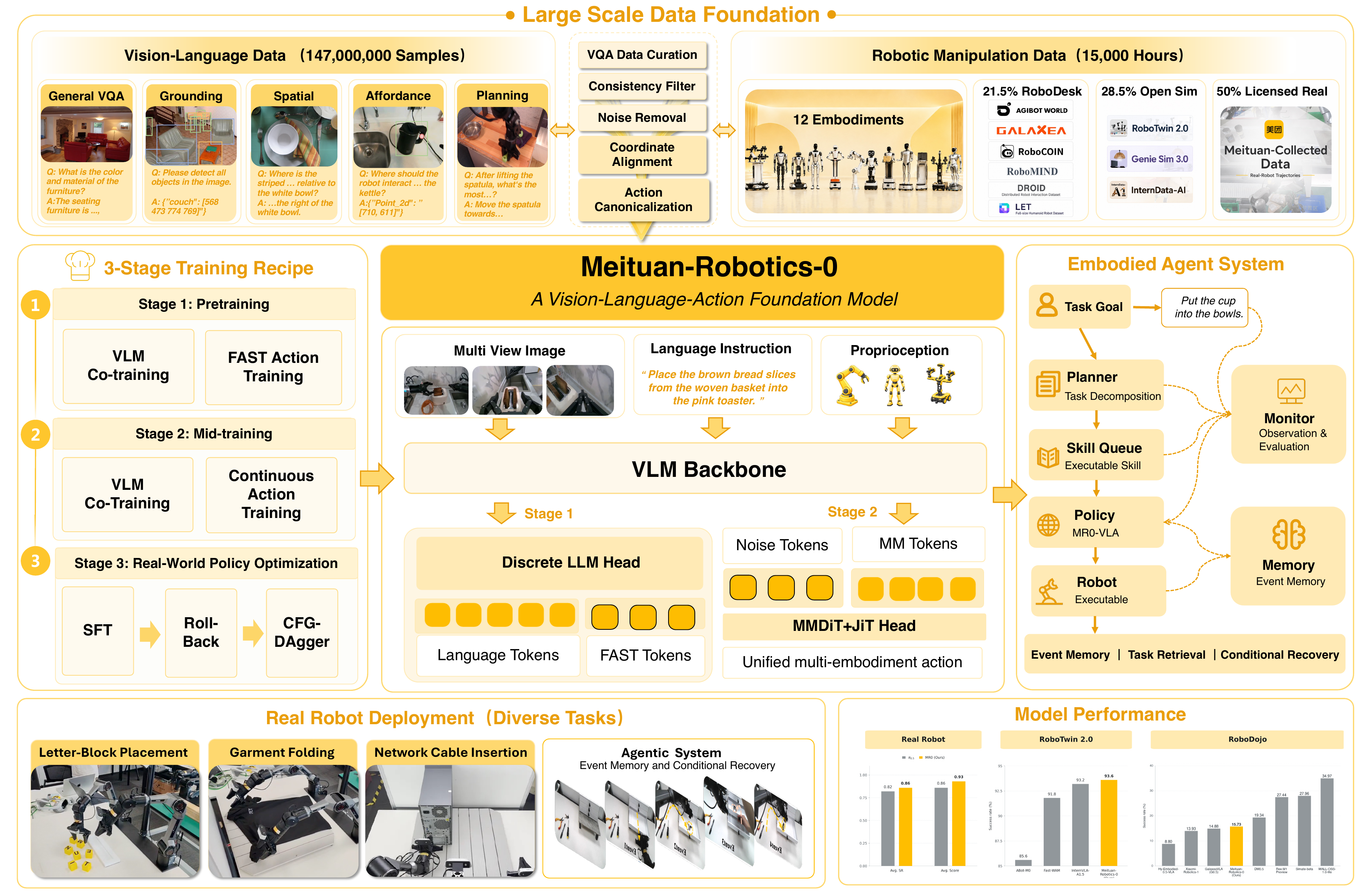}
\caption{\textbf{Overview of Meituan-Robotics-0.} The framework integrates large-scale multimodal data, three-stage VLA training, an event-driven embodied-agent system, and real-robot deployment. The RoboDojo panel orders the displayed VLA models in ascending order of average success rate.}
\vspace{-5mm}
\label{fig:mr0_overview}
\end{figure}

We present \textbf{Meituan-Robotics-0 (MR0)}, a desktop-manipulation framework built around a large-scale data foundation, a generalist VLA policy, and an event-driven embodied agent, as summarized in~\Cref{fig:mr0_overview}. The data foundation combines 147 million vision--language samples with 15,060 hours of curated robot trajectories spanning 12 embodiments. The vision--language corpus provides general visual-semantic knowledge together with embodied supervision for grounding, scene understanding, affordance prediction, and task reasoning. 
The robot corpus comprises three complementary components: 
\textbf{RoboDesk}, a curated collection of open-source real-robot
trajectories totaling 3,235 hours; 7,530 hours of
licensed real-robot data spanning seven robot configurations; and
4,295 hours of open simulation data. The complete corpus is
processed through temporal and visual filtering, kinematic-consistency
checks, cross-embodiment action canonicalization, and hierarchical task
alignment. RoboDesk and the simulation data broaden task, environment,
and embodiment coverage, while the licensed real-robot data provide
task-focused manipulation experience.

Building on this aligned data foundation, the MR0 policy follows a three-stage training recipe that separates the acquisition of broad action priors from continuous-control learning and real-world adaptation. A shared multimodal backbone is first paired with a discrete action head and pre-trained jointly on vision-language data and tokenized robot actions. Continuous mid-training then introduces an MMDiT~\cite{esser2024mmdit} action head and optimizes flow matching over a unified end-effector representation, transferring the pre-trained features to fine-grained control across single-arm and bimanual embodiments. Real-world post-training~\cite{li2026simplevla,li2025vla,ye2025vla} starts with supervised finetuning on successful demonstrations. HG-DAgger~\cite{kelly2019hgdagger} adds human corrections from policy-induced error states, while RECAP (RL with Experience and Corrections via Advantage-conditioned Policies)~\cite{amin2025pi06star} derives advantage labels from outcome-labeled autonomous rollouts. We further introduce Rollback, a collection mechanism that returns to a historical anchor to collect a human replacement while retaining the original policy segment. These paired segments supply direct positive and negative labels to CFG-DAgger and remove the need for value learning and advantage labeling. At deployment, asynchronous action-chunk merging integrates new predictions with commands already being executed, maintaining smooth control without blocking on policy inference.

The VLA policy does not explicitly maintain task state across subtasks. MR0 therefore adds an event-driven embodied agent~\cite{huo2026abot,li2026roboclaw,zhang2026harness} to retain relevant context, track task progress, and revise plans as conditions change. A Monitor summarizes short visual windows into task-relevant events, and a Memory Manager consolidates these events into a persistent record of task progress and environmental conditions. The Planner updates the active subtask only when a new event changes task progress, reveals a failure, or alters whether the current objective is feasible. Motor execution consequently remains with the VLA policy, while the Planner is invoked only when task-level reassessment is needed. This separation enables the system to retrieve information acquired in earlier tasks, suspend an objective whose prerequisites are temporarily unavailable, and resume it after the required conditions are restored.

We evaluate the policy and agent under protocols matched to their respective roles. MR0 achieves 93.6\% average success on RoboTwin~2.0~\cite{robotwin2}. On RoboDojo~\cite{chen2026robodojo}, the history-augmented policy achieves an average success rate of 15.73\%. 
With approximately 13.6k hours of embodied training data, MR0 achieves
competitive performance across the evaluated manipulation capabilities.
Human-intervention data improve both MR0 and $\pi_{0.5}$ across three real-robot tasks.
Under matched HG-DAgger post-training, MR0 achieves higher average success rate and progress score than $\pi_{0.5}$ across the three tasks.
In a separate comparison on Ethernet insertion, MR0 trained with CFG-DAgger reaches 70\% success and a progress score of 0.90, outperforming HG-DAgger and RECAP on the same Rollback data.
Qualitative agent deployments further illustrate cross-task event retrieval and task resumption following temporary interruptions.
Together, these evaluations show how aligned data, continuous policy learning, real-world post-training, and event-level memory address complementary requirements of general-purpose desktop manipulation.

%% file: sections/02-data_curation.tex
\section{Data Curation}
\label{sec:data}

\input{tables/embodied-vqa-ablation}

To equip MR0 with both broad visual-semantic understanding and diverse
cross-embodiment manipulation experience, we independently curate two
complementary data resources: 147 million vision--language samples and
15,060 hours of robot trajectories spanning 12 embodiments. The
vision--language corpus provides supervision for general perception and
embodied reasoning. Constructed through a separate pipeline, the robot
corpus combines RoboDesk, licensed real-robot data, and open simulation
data for action learning across heterogeneous hardware. We describe the
two resources in Secs.~\ref{sec:vlm_data}
and~\ref{sec:vla_data}, respectively.

\subsection{Vision-Language Data}
\label{sec:vlm_data}

\begin{figure}[t]
    \centering
    \includegraphics[width=0.9\linewidth]{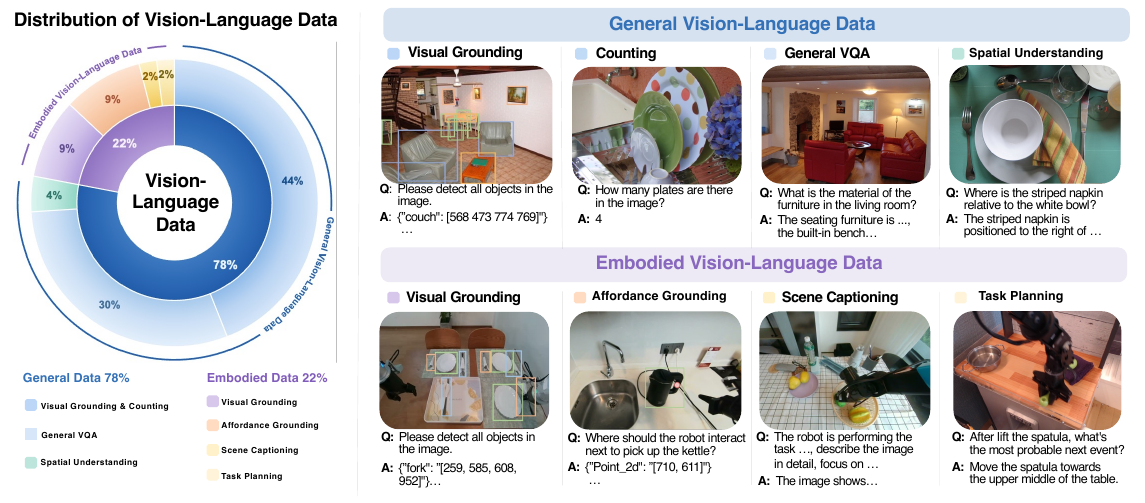}
    \vspace{2pt}
    \caption{
    \textbf{Vision-language dataset.}
    The corpus contains approximately 147 million samples before
    training-time resampling, comprising 78\% general and 22\% embodied
    vision-language data.
    Representative examples are shown on the right.
    }
    \label{fig:vqa_data}
    \vspace{-15pt}
\end{figure}

Visual reasoning for manipulation requires both broad perceptual knowledge and embodied understanding grounded in robot interaction. As shown in~\Cref{fig:vqa_data}, we organize the 147M vision-language corpus into two complementary subsets. General-purpose data account for 78\% and provide supervision for visual question answering, spatial understanding, visual grounding, and counting. The remaining 22\% consist of embodied data for scene captioning, task planning, visual grounding, and affordance grounding. Affordance grounding jointly localizes the manipulated object with a bounding box and identifies a 2D interaction point at which the robot should act. These corpus statistics are computed before training-time resampling.

For the embodied VQA component, we augment the collected open-source datasets with a large-scale in-house corpus derived from real-robot trajectories. To determine which supervision types should be prioritized in this corpus, we conduct controlled CALVIN ABC-to-D ablations over eight candidate tasks and use the results to guide subsequent large-scale data construction, as shown in ~\Cref{tab:embodied-vqa-ablation}. The candidates span three capability families: spatial localization, action-related understanding, and semantic and task reasoning. All variants use the same StarVLA~\citep{community2026starvla} configuration, are trained on environments A, B, and C, and are evaluated on environment D. \Cref{tab:embodied-vqa-ablation} reports success rates for completing sequences of 1 to 5 consecutive subtasks, along with the average completed sequence length. The results support including 2D target-affordance grounding and scene captioning as the action-related and semantic supervision types, respectively. For spatial localization, we choose 2D grounding because it avoids costly geometric annotations and scales more readily than 3D grounding. Taken together, the empirical results and practical annotation considerations lead us to focus our in-house data expansion on three types: 2D target-affordance grounding, scene captioning, and 2D grounding.

To ensure consistent and scalable data construction, we develop task-specific annotation pipelines for the selected supervision types. For 2D grounding, we apply Set-of-Mark prompting~\citep{yang2023setofmark}: the Segment Anything Model (SAM)~\citep{kirillov2023segany} proposes and marks candidate regions, and a vision-language model assigns an object identity to each region to form grounding question--answer pairs. For 2D affordance grounding, we obtain the target bounding box and interaction point from the key interaction frame of each action phase and propagate both annotations to earlier frames within that phase. For scene captioning, a vision-language model conditions on the robot observation and task context to describe task-relevant objects, spatial relations, gripper state, and execution progress.

\begin{wrapfigure}{r}{0.4\textwidth}
    \vspace{-20pt}
    \centering
    \includegraphics[
        width=\linewidth
    ]{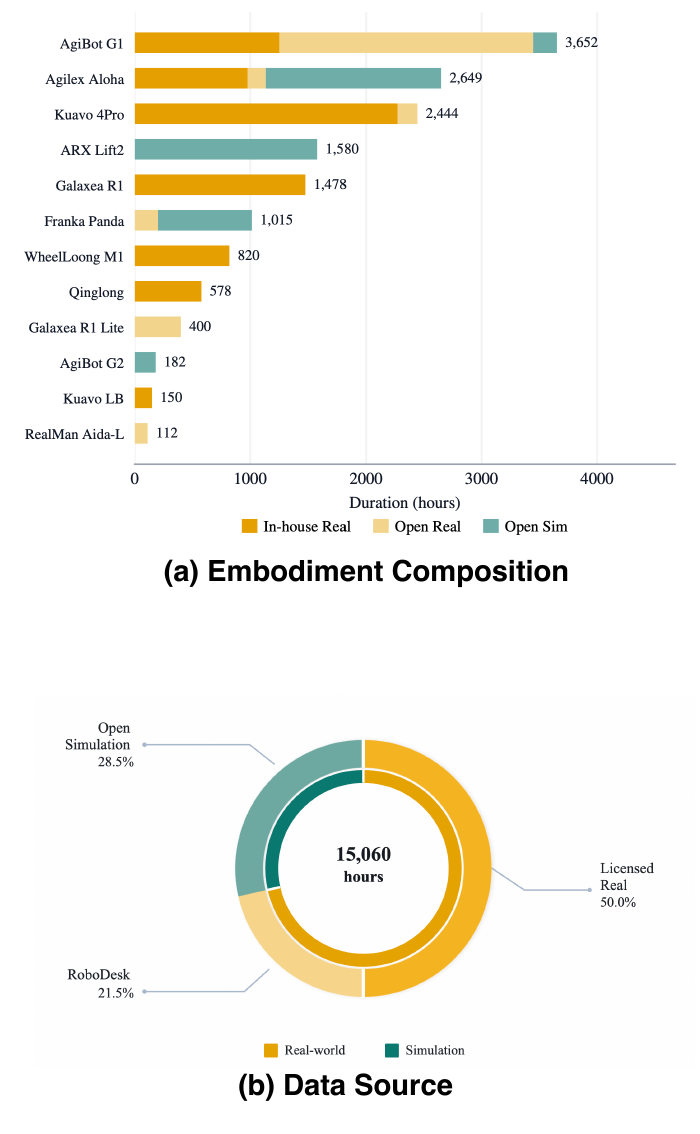}
    \caption{\textbf{Composition of the robot corpus.}
The 15,060-hour corpus spans 12 embodiments and comprises RoboDesk,
a curated collection of open-source real-robot trajectories totaling
3,235 hours (21.5\%); 7,530 hours (50.0\%) of licensed real-robot data;
and 4,295 hours (28.5\%) of open simulation data.}
    \label{fig:robot-data-source}
    \vspace{-55pt}
\end{wrapfigure}

\subsection{Robotic Data}
\label{sec:vla_data}


To provide diverse action supervision for robot control, we assemble
15,060 hours of tabletop trajectories spanning 12 embodiments. We
restrict the corpus to stationary-base manipulation and exclude segments
containing base motion. As summarized in ~\Cref{fig:robot-data-source}, the corpus consists of RoboDesk (3,235 hours; 21.5\%), licensed real-robot data (7,530 hours; 50.0\%),
and open simulation data (4,295 hours; 28.5\%). RoboDesk denotes the six
open-source real-world datasets retained after curation, which are listed
together with the simulation sources in ~\Cref{tab:open-robot-datasets}.

\subsubsection{Data Curation Pipeline.} 

\begin{figure}[t]
    \centering
    \includegraphics[width=\linewidth]{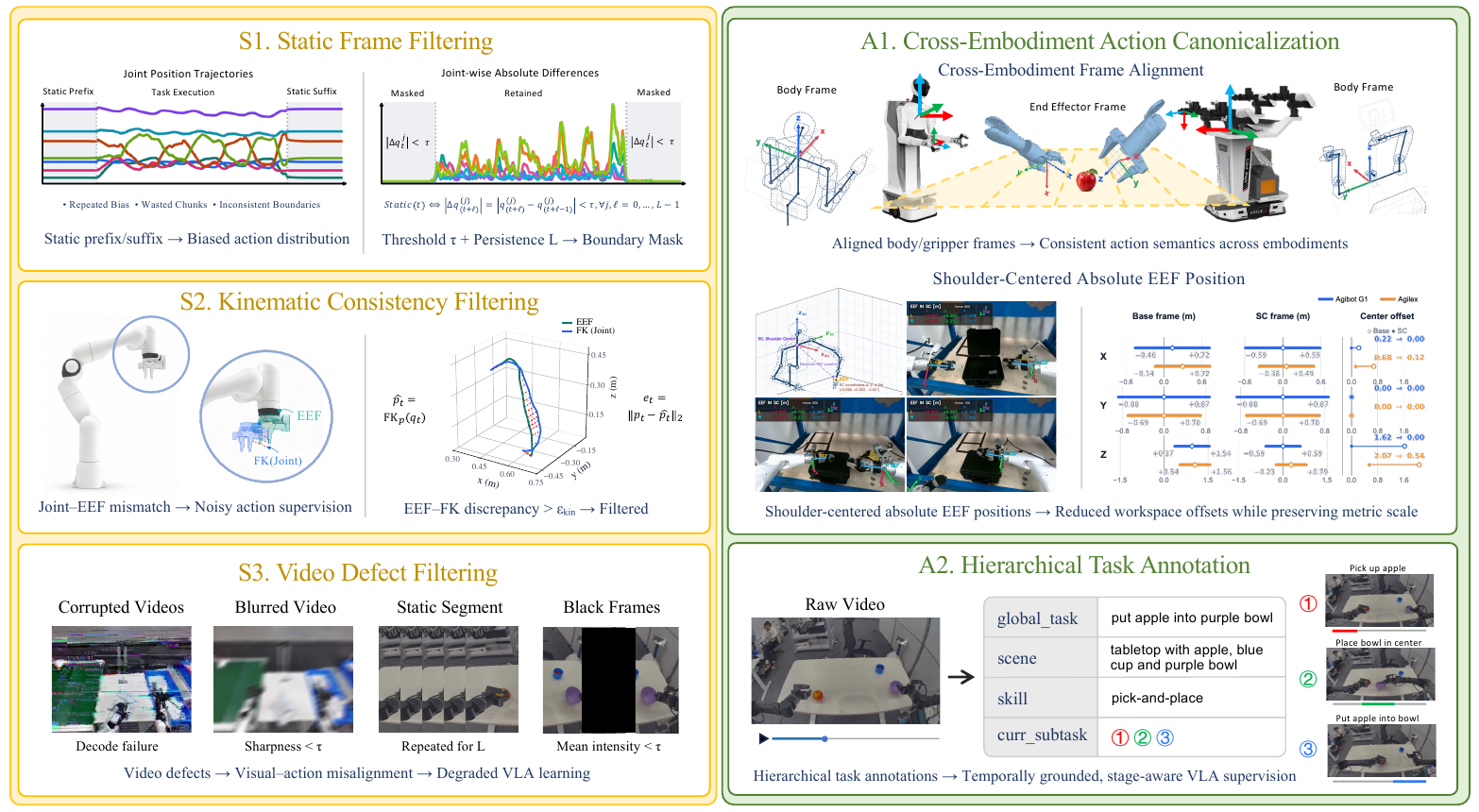}
    \caption{\textbf{Robot trajectory curation and alignment pipeline.} S1--S3 remove static boundaries, kinematic inconsistencies, and video defects; A1 canonicalizes actions across embodiments, and A2 aligns multi-granularity task annotations with the retained trajectories.}
    \vspace{-15pt}
    \label{fig:robot-data-curation}
\end{figure}

Raw trajectories contain idle boundaries, inconsistent kinematic records, visual defects, embodiment-specific action conventions, and uneven language annotations. We address these issues with three filtering stages (S1--S3) followed by two alignment stages (A1--A2). S1 removes static prefixes and suffixes, S2 rejects inconsistent joint--EEF recordings, and S3 removes corrupted visual segments. A1 canonicalizes actions across embodiments, while A2 aligns multi-granularity task annotations with the retained trajectories. \Cref{fig:robot-data-curation} summarizes the pipeline.

\textbf{Static-boundary filtering.} Teleoperated recordings often begin before manipulation and continue through task completion or reset, producing long intervals of near-zero motion at episode boundaries. Let $q_t \in \mathbb{R}^J$ be the recorded joint vector and $\Delta q_t^{(j)} = q_t^{(j)} - q_{t-1}^{(j)}$. A time index is considered static only when every joint remains below a motion threshold for $L$ consecutive steps:

\begin{equation}
\begin{aligned}
\operatorname{Static}(t)
&\Longleftrightarrow
\left|\Delta q_{t+\ell}^{(j)}\right| < \tau_{\mathrm{stat}}, \\
&\qquad
\forall j \in \{1,\ldots,J\},\quad
\forall \ell \in \{0,\ldots,L-1\}.
\end{aligned}
\label{eq:static-boundary-filter}
\end{equation}

The persistence window prevents brief low-speed motions from being classified as idle. We trim only static runs connected to an episode boundary, preserving deliberate pauses during task execution. The corresponding video, state, and action intervals are removed jointly to maintain temporal synchronization.

\textbf{Kinematic-consistency filtering.} For each embodiment, we reconstruct the end-effector position from the recorded joints using the corresponding URDF and compare it with the logged EEF position in a common frame. Samples exceeding an embodiment-specific error threshold are masked, and trajectories with persistent violations are discarded.

\input{tables/data_source}

\textbf{Video-defect filtering.} We detect decoding failures, blur, frozen segments, and black frames. Corrupted intervals are removed from all synchronized streams, while episodes with persistent defects are discarded. Detection thresholds are calibrated separately by embodiment and data source.

\textbf{Cross-embodiment action canonicalization.} To make action channels physically consistent across embodiments, we define common right-handed body and gripper frames. The body axes point forward, left, and up. At the open-gripper zero pose, $+z$ points from the mounting flange to the gripper tip, $+y$ from the right finger to the left, and $+x$ completes the frame. EEF poses are expressed in a canonical shoulder-centered frame, denoted $\mathrm{SC}$. The frame is shared by both arms of a bimanual robot; for a single-arm embodiment without an explicit shoulder center, the arm-root frame is used instead. For arm $i$, the action contains the absolute EEF position ${}^{\mathrm{SC}}p_{\mathrm{EEF},t}^{(i)} \in \mathbb{R}^3$, a continuous 6D orientation representation $r_t^{(i)} \in \mathbb{R}^6$~\citep{zhou2019continuity}, and normalized gripper closure $g_t^{(i)} \in [0,1]$, where 0 and 1 denote fully open and fully closed, respectively:
\begin{equation}
a_t^{(i)} =
\left[
\left({}^{\mathrm{SC}}p_{\mathrm{EEF},t}^{(i)}\right),\,
\left(r_t^{(i)}\right),\,
g_t^{(i)}
\right]^{\!\top}
\in\mathbb{R}^{10},
\qquad
a_t =
\left[
\left(a_t^{(\mathrm{L})}\right),\,
\left(a_t^{(\mathrm{R})}\right)
\right]^{\!\top}
\in\mathbb{R}^{20}.
\label{eq:canonical-action-vector}
\end{equation}

Single-arm samples use the 10-dimensional form. Let $\mathrm{B}$ be the source body frame, ${}^{\mathrm{SC}}T_{\mathrm{B}}$ the embodiment-specific rigid transform from $\mathrm{B}$ to $\mathrm{SC}$, and ${}^{\mathrm{B}}T_{\mathrm{EEF},t}^{(i)}$ the source EEF target pose. Canonicalization is
\begin{equation}
{}^{\mathrm{SC}}T_{\mathrm{EEF},t}^{(i)}
=
{}^{\mathrm{SC}}T_{\mathrm{B}}\,{}^{\mathrm{B}}T_{\mathrm{EEF},t}^{(i)},
\qquad
r_t^{(i)} = \rho_{6\mathrm{D}}\!\left({}^{\mathrm{SC}}R_{\mathrm{EEF},t}^{(i)}\right).
\label{eq:action-canonicalization}
\end{equation}

The rotation and translation blocks of ${}^{\mathrm{SC}}T_{\mathrm{EEF},t}^{(i)}$ provide the canonical orientation and metric absolute position target, respectively; the translation is not a temporal displacement.

\textbf{Hierarchical task annotation.} Each episode receives an overall goal in \texttt{global\_task}, while \texttt{curr\_subtask} labels the interval in which each execution stage is active. The auxiliary \texttt{scene} field describes task-relevant objects and their spatial arrangement, and \texttt{skill} names the manipulation primitive. The resulting corpus pairs synchronized observations and canonical EEF actions with episode-level goals, time-aligned subtask labels, and scene and skill context, supporting mixed-granularity instruction conditioning during training.

%% file: tables/embodied-vqa-ablation.tex
\begin{wraptable}{r}{0.49\textwidth}
    \vspace{-10pt}
    \centering

    \captionsetup{
        font=small,
        skip=4pt
    }

    \caption{
    \textbf{Ablation of embodied vision--language supervision in the CALVIN ABC$\rightarrow$D setting.} Bold indicates the highest Avg.\ Len.\ within each capability family.
    }
        
    \label{tab:embodied-vqa-ablation}

    \scriptsize
    \setlength{\tabcolsep}{1.5pt}
    \renewcommand{\arraystretch}{1.05}

    \resizebox{\linewidth}{!}{%
        \begin{tabular}{@{}llcccccc@{}}
            \toprule

            \makecell[l]{
                \textbf{Capability}\\
                \textbf{family}
            }
            &
            \makecell[l]{
                \textbf{Training}\\
                \textbf{setting}
            }
            &
            \makecell[c]{
                \textbf{Avg.}\\
                \textbf{Len.}
            }
            &
            \makecell[c]{
                \textbf{Task}\\
                \textbf{1}
            }
            &
            \makecell[c]{
                \textbf{Task}\\
                \textbf{2}
            }
            &
            \makecell[c]{
                \textbf{Task}\\
                \textbf{3}
            }
            &
            \makecell[c]{
                \textbf{Task}\\
                \textbf{4}
            }
            &
            \makecell[c]{
                \textbf{Task}\\
                \textbf{5}
            }
            \\
            \midrule

            Baseline
            & --
            & 3.832
            & 91.0\%
            & 83.0\%
            & 76.2\%
            & 70.2\%
            & 62.8\%
            \\
            \midrule

            \multirow{3}{*}{
                \makecell[l]{
                    Spatial\\
                    localization
                }
            }
            & \(+\) 2D Grounding
            & 3.941
            & 91.3\%
            & 83.7\%
            & 78.1\%
            & 73.2\%
            & 67.8\%
            \\
            \cmidrule(lr){2-8}

            & \(+\) 2D Pointing
            & 3.806
            & 90.8\%
            & 81.8\%
            & 74.8\%
            & 69.9\%
            & 63.3\%
            \\
            \cmidrule(lr){2-8}

            & \(+\) 3D Grounding
            & \textbf{4.037}
            & 92.4\%
            & 85.3\%
            & 79.6\%
            & 75.7\%
            & 70.7\%
            \\
            \midrule

            \multirow{3}{*}{
                \makecell[l]{
                    Action-related\\
                    understanding
                }
            }
            &
            \makecell[l]{
                \(+\) 2D Target-Affordance\\
                Grounding
            }
            & \textbf{4.102}
            & 93.3\%
            & 87.9\%
            & 82.5\%
            & 76.4\%
            & 70.1\%
            \\
            \cmidrule(lr){2-8}

            &
            \makecell[l]{
                \(+\) 3D Target-Affordance\\
                Grounding
            }
            & 4.098
            & 93.7\%
            & 89.9\%
            & 81.2\%
            & 73.9\%
            & 71.1\%
            \\
            \cmidrule(lr){2-8}

            & \(+\) 2D Trajectory
            & 3.987
            & 92.0\%
            & 85.0\%
            & 79.2\%
            & 74.0\%
            & 68.5\%
            \\
            \midrule

            \multirow{2}{*}{
                \makecell[l]{
                    Semantic and\\
                    task reasoning
                }
            }
            & \(+\) Subtask Prediction
            & 3.851
            & 90.0\%
            & 82.2\%
            & 76.2\%
            & 71.7\%
            & 65.0\%
            \\
            \cmidrule(lr){2-8}

            & \(+\) Scene Captioning
            & \textbf{4.082}
            & 93.1\%
            & 87.1\%
            & 81.4\%
            & 76.2\%
            & 70.4\%
            \\
            \bottomrule
        \end{tabular}%
    }

    \vspace{-35pt}
\end{wraptable}

%% file: tables/data_source.tex
\begin{wraptable}{r}{0.4\textwidth}
    \centering
    \vspace{-4mm}
    \captionsetup{
        font=small,
        skip=2pt
    }
    \caption{
        Open-source datasets retained after curation.
    }
    \label{tab:open-robot-datasets}

    \footnotesize
    \setlength{\tabcolsep}{3pt}
    \renewcommand{\arraystretch}{1}
    \begin{tabularx}{\linewidth}{@{}Xrr@{}}
        \toprule

        \textbf{Data source}
        & \textbf{Hours}
        & \textbf{Share} \\
        \midrule

        \multicolumn{3}{c}{
            \textbf{RoboDesk (real-world): 3,235 h (21.5\%)}
        } \\
        \cmidrule(lr){1-3}

        AgiBotWorld-Beta~\citep{agibotworld}
        & 2,174
        & 14.4\% \\

        Galaxea Open-World~\citep{galaxea}
        & 333
        & 2.2\% \\

        RoboCOIN~\citep{robocoin2025}
        & 311
        & 2.1\% \\

        LET~\citep{lejurobotics2025let}
        & 172
        & 1.1\% \\

        DROID~\citep{khazatsky2024droid}
        & 164
        & 1.1\% \\

        RoboMIND~\citep{robomind}
        & 81
        & 0.5\% \\

        \midrule
        \multicolumn{3}{c}{
            \textbf{Simulation: 4,295 h (28.5\%)}
        } \\

        \midrule

        InternData-A1~\citep{tian2025interndata}
        & 3,967
        & 26.3\% \\

        Genie Sim 3.0~\citep{geniesim3}
        & 213
        & 1.4\% \\

        RoboTwin 2.0~\citep{robotwin2}
        & 115
        & 0.8\% \\

        \midrule
        \multicolumn{3}{c}{
            \makecell[c]{
                \textbf{Open-source total: 7,530 h (50\%)}
            }
        } \\

        \bottomrule
    \end{tabularx}
    \vspace{-15pt}
\end{wraptable}

%% file: sections/03-model.tex
\section{Meituan-Robotics-0}
\label{sec:model}

Meituan-Robotics-0 aims to transfer broad priors learned from heterogeneous multimodal data to precise and robust real-robot control. This transition requires the model to preserve its visual-semantic representations while adapting its action interface and supervision to continuous cross-embodiment control and policy-induced error states. We first describe the policy architecture in~\Cref{sec:policy-architecture} and then present the three-stage training pipeline in~\Cref{sec:progressive-training}: discrete multimodal pre-training, continuous cross-embodiment mid-training, and real-world policy optimization.

\subsection{Policy Architecture}
\label{sec:policy-architecture}

\begin{wrapfigure}{r}{0.4\columnwidth}
    \vspace{-15pt}
    \centering
    \includegraphics[
        width=\linewidth,
        keepaspectratio
    ]{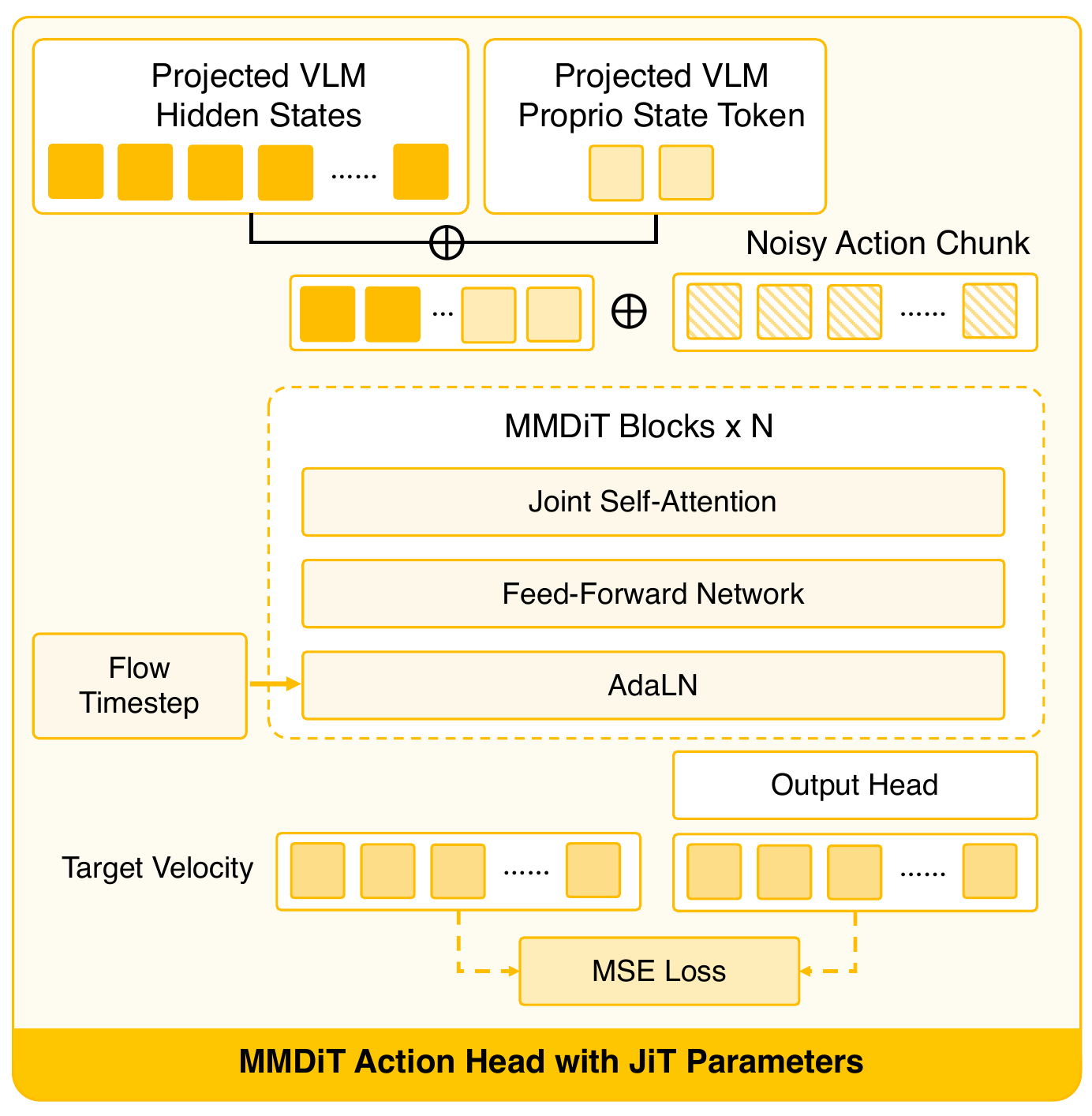}
    \caption{
        \textbf{Continuous action head.} MMDiT denoises action chunks through flow matching.
    }
    \label{fig:mmdit}
    \vspace{-3pt}
\end{wrapfigure}

MR0 formulates low-level desktop manipulation as reactive action-chunk prediction from the current observation. At this level, the policy maintains no temporal history and conditions only on the inputs available at the current decision step. Given multi-view observations $o_t=\{I_t^{(c)}\}_{c=1}^{C}$, language instruction $l$, and proprioceptive state $s_t$ at time $t$, MR0 predicts an $H$-step action chunk
\begin{equation}
\pi_\theta\!\left(a_{t:t+H-1}\mid o_t,l,s_t\right).
\label{eq:policy-formulation}
\end{equation}

Following the common VLA design~\citep{qwenvla2026,black2024pi0,nvidia2025gr00t}, MR0 realizes this mapping with a shared vision--language backbone and two stage-specific output heads, as summarized in~\Cref{fig:mr0_overview}. The shared backbone is initialized from Qwen3.5-4B~\citep{qwen35-2026} and remains trainable throughout the pipeline. During pre-training, FAST~\citep{pertsch2025fast} applies a discrete cosine transform along the temporal axis and compresses the resulting coefficients with byte-pair encoding. The encoded action chunks share the backbone vocabulary and are predicted by the same autoregressive token head used for text.

For mid-training and all downstream stages, MR0 replaces the discrete interface with a continuous MMDiT action head~\citep{esser2024mmdit,black2024pi0}. As illustrated in~\Cref{fig:mmdit}, the head jointly processes an observation stream containing projected backbone features and a proprioceptive-state token, and an action stream containing a noisy action chunk. Joint self-attention exchanges information between the two streams, while the flow timestep modulates block normalization through AdaLN. The state token remains on the observation side, providing condition-side proprioceptive context throughout action generation.


\subsection{Progressive Training}
\label{sec:progressive-training}

MR0 is trained progressively to separate three sources of difficulty: heterogeneous multimodal learning, continuous cross-embodiment control, and recovery from policy-induced errors.

\subsubsection{Discrete Multimodal Pre-Training}
\label{sec:discrete-pretraining}

The shared token interface allows robot trajectories and vision--language samples to use the same autoregressive objective without a separate action decoder. FAST action tokens occupy a dedicated vocabulary block whose embeddings are initialized with the mean of the original embedding table. The remaining backbone parameters are inherited and trained end-to-end, and all camera views share the native vision encoder and aspect-preserving smart-resize policy.

Each optimization step draws one batch from the VLA action pool and one from the vision--language pool. The joint objective is
\begin{equation}
\mathcal{L}_{\mathrm{pre}}
=
\lambda_{\mathrm{vla}}\mathcal{L}_{\mathrm{action}}
+
\lambda_{\mathrm{vlm}}\mathcal{L}_{\mathrm{vqa}},
\qquad
\lambda_{\mathrm{vla}}=1.0,
\quad
\lambda_{\mathrm{vlm}}=0.1,
\label{eq:pretraining-objective}
\end{equation}

where both terms are next-token cross-entropy over their respective targets. The action loss grounds the backbone in robot behavior, while the VQA loss preserves perceptual and semantic supervision. We train the backbone and extended action-token embeddings end to end for 500{,}000 steps on 256 GPUs using AdamW with cosine learning-rate decay. The resulting checkpoint initializes mid-training. Because FAST tokenization~\citep{pertsch2025fast} introduces action-reconstruction error, downstream stages discard the token interface in favor of continuous prediction.

\subsubsection{Continuous Cross-Embodiment Mid-Training}
\label{sec:continuous-midtraining}

The second stage retains the token-pre-trained backbone and initializes the MMDiT action head~\citep{esser2024mmdit} from scratch. This transfers the visual-semantic and behavioral representation learned during pre-training while replacing quantized action generation with continuous control.

\textbf{Continuous action objective.} We train the MMDiT head with the JiT clean-sample parameterization~\citep{li2026back}. Following flow matching~\citep{lipman2023flow}, a clean chunk $a_0$ is perturbed into a noisy sample $a_\tau$, from which the head predicts $\hat a_0$ rather than velocity directly. Prediction and target are converted to velocity form using a shared clamped denominator $d_\tau$ with $t_\epsilon=0.05$, and the model is optimized by mean squared error.

\textbf{Unified action representation.} To apply the continuous objective across heterogeneous embodiments, MR0 maps each native action to the canonical end-effector convention defined in~\Cref{sec:vla_data}. Each action chunk has a fixed width $d_{\max}=20$, corresponding to two 10-dimensional end-effector actions, while a binary mask $m\in\{0,1\}^{d_{\max}}$ identifies the valid channels for each embodiment:
\begin{equation}
\mathcal{L}_{\mathrm{masked}}
=
\sum_{h=1}^{H}\sum_{j=1}^{d_{\max}}m_j\ell_{h,j}
\;\Big/\;
H\sum_{j=1}^{d_{\max}}m_j,
\label{eq:masked-action-loss}
\end{equation}
where $\ell_{h,j}$ is the element-wise JiT loss. Padded channels produce no gradient. For a downstream embodiment with a different action width, only the dimension-dependent projections are reinitialized.

In this stage, we train on 128 GPUs using AdamW with cosine learning-rate decay, applying a lower learning rate to the transferred backbone and a higher rate to the newly initialized action head. The training mixture starts from the data pools summarized in~\Cref{fig:robot-data-source}. We rebalance it by removing a 340-million-frame synthetic sub-corpus, retaining 10\% of each of the two largest general-VQA sources, and sampling VLA datasets in proportion to the square root of their frame counts. To expose the model to different levels of linguistic abstraction, we sample the local subtask label or the episode-level goal with equal probability when both are available and otherwise use the available annotation. This \texttt{langmix} strategy pairs the same actions with both fine-grained and task-level instructions.

\subsubsection{Real-World Policy Optimization}
\label{sec:real-world-policy-optimization}

We first fine-tune the mid-training checkpoint $\pi_{\mathrm{mid}}$ on successful demonstrations to obtain an initial task policy. To address policy-induced distribution shift~\citep{ross2011reduction}, we collect human-in-the-loop (HIL) corrections and compare three post-training algorithms. Human-gated DAgger~\citep{kelly2019hgdagger} imitates demonstrations and corrections; RECAP~\citep{amin2025pi06star} uses trajectory outcomes to learn advantage-conditioned behavior; and CFG-DAgger uses positive and negative segments supplied by Rollback. Each collection round adds to the cumulative dataset, and each policy is trained anew from $\pi_{\mathrm{mid}}$. All three methods execute action chunks through the asynchronous controller in~\Cref{sec:asynchronous_policy_execution}.

\textbf{Human-gated DAgger.} The policy controls the robot until an operator intervenes to correct a stall, error, or unsafe motion. Only human-controlled segments are retained as corrective supervision. Demonstrations cover the task procedure, while HIL corrections add recovery behavior in states encountered by the policy. Details are in~\Cref{sec:real_robot_dagger}.

\begin{figure}[t]
    \centering
    \includegraphics[width=0.8\linewidth]{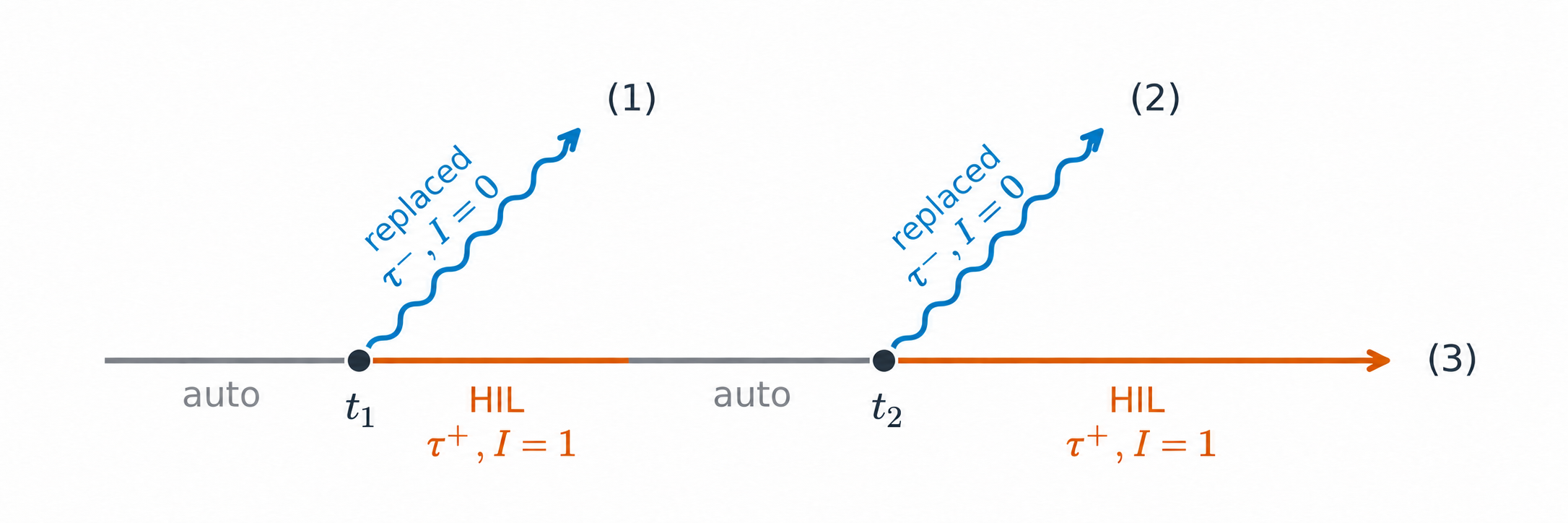}
    \caption{\textbf{Rollback data collection.} Dots mark the rollback anchors $t_1$ and $t_2$. Wavy blue branches (1) and (2) are replaced policy segments ($\tau^-$, $I=0$); solid orange HIL segments are their human replacements ($\tau^+$, $I=1$), with corrected execution continuing toward (3). These conditions are used by CFG-DAgger. Gray segments denote other autonomous execution.}
    \label{fig:rollback_collection}
\end{figure}

\textbf{Rollback data collection.} We introduce Rollback to pair replaced policy segments with human corrections from a shared anchor. An operator pauses execution near failure, selects an earlier recorded state, returns the robot to it using action history, and supplies a corrective continuation. In~\Cref{fig:rollback_collection}, each anchor ($t_1$ or $t_2$) pairs a replaced segment $\tau^-$ with its human replacement $\tau^+$. Alongside demonstrations and human corrections, Rollback retains suboptimal segments to inform value learning and guide policy updates.

\textbf{RECAP policy optimization.} 
RECAP trains its value model on both successful and failed trajectories
reconstructed from Rollback episodes. A trajectory ending with a replaced
policy segment $\tau^-$ is assigned a failure label. Policy training uses
only successful trajectories.\footnote{We intentionally exclude failed
trajectories from policy training because this yielded better performance
in our setting.}
The learned value model provides $N$-step advantages for assigning positive ($I=1$) or base ($I=\varnothing$) conditions to policy-training chunks. Both branches share the flow-matching loss. At inference, action-space classifier-free guidance (CFG) combines base and positive predictions generated from the same initial noise. Details are in~\Cref{sec:recap_details}.

\textbf{CFG-DAgger conditional policy training.} CFG-DAgger directly trains the policy on demonstrations and $\tau^+$ as positive data and $\tau^-$ as negative data. A shared policy learns positive ($I=1$), negative ($I=0$), and unconditional ($I=\varnothing$) branches with the same flow-matching loss. At inference, CFG combines positive and negative predictions to favor corrective behavior and steer away from replaced actions. Details are in~\Cref{sec:cfg_dagger_details}.

By obtaining labels directly through Rollback collection, CFG-DAgger avoids value learning and advantage labeling, while outperforming RECAP in our real-robot evaluation (\Cref{sec:real_robot_policy_evaluation}).

%% file: sections/04-deployment.tex
\section{Embodied Manipulation System}
\label{sec:deployment-agent-system}

\begin{figure}[t]
    \centering
\includegraphics[width=0.95\linewidth]{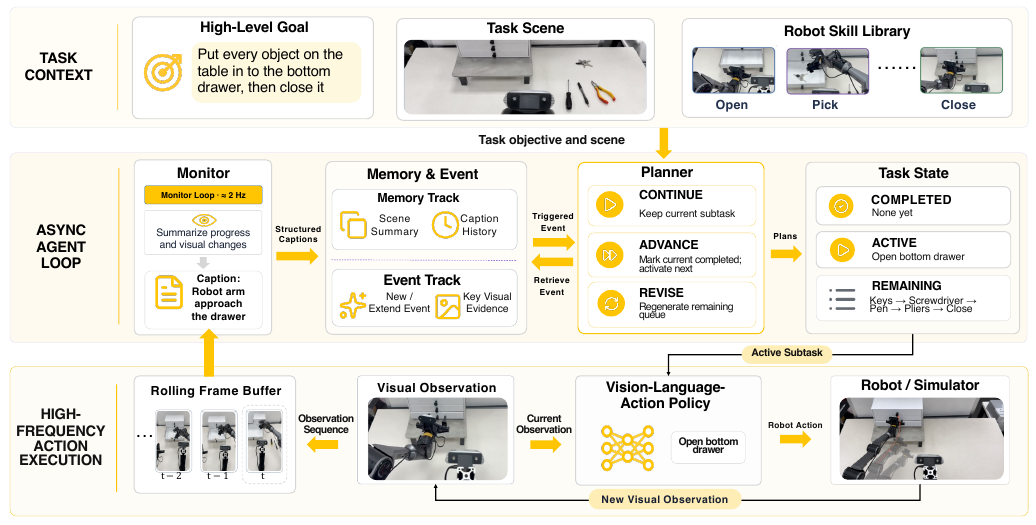}
    \caption{\textbf{Event-driven embodied-agent framework.} The asynchronous
    agent loop converts visual observations into structured captions,
    consolidates them into events, and triggers the Planner when task-state
    reassessment is required; the active subtask conditions the high-frequency
    VLA control loop.}
    \label{fig:agent_framework}
    \vspace{-12px}
\end{figure}

MR0 augments the trained VLA policy with an inference-time system for continuous control and long-horizon task execution. An asynchronous execution layer aligns variable-latency action predictions with robot control, as detailed in~\Cref{sec:asynchronous_policy_execution}. At a higher level, an event-driven agent retains task-relevant state beyond the policy's current observation. As summarized in~\Cref{fig:agent_framework}, motor control remains with the VLA policy, and high-level planning is activated only by events that alter task progress or feasibility.

\begin{figure}[t]
    \centering
\includegraphics[width=0.75\linewidth]{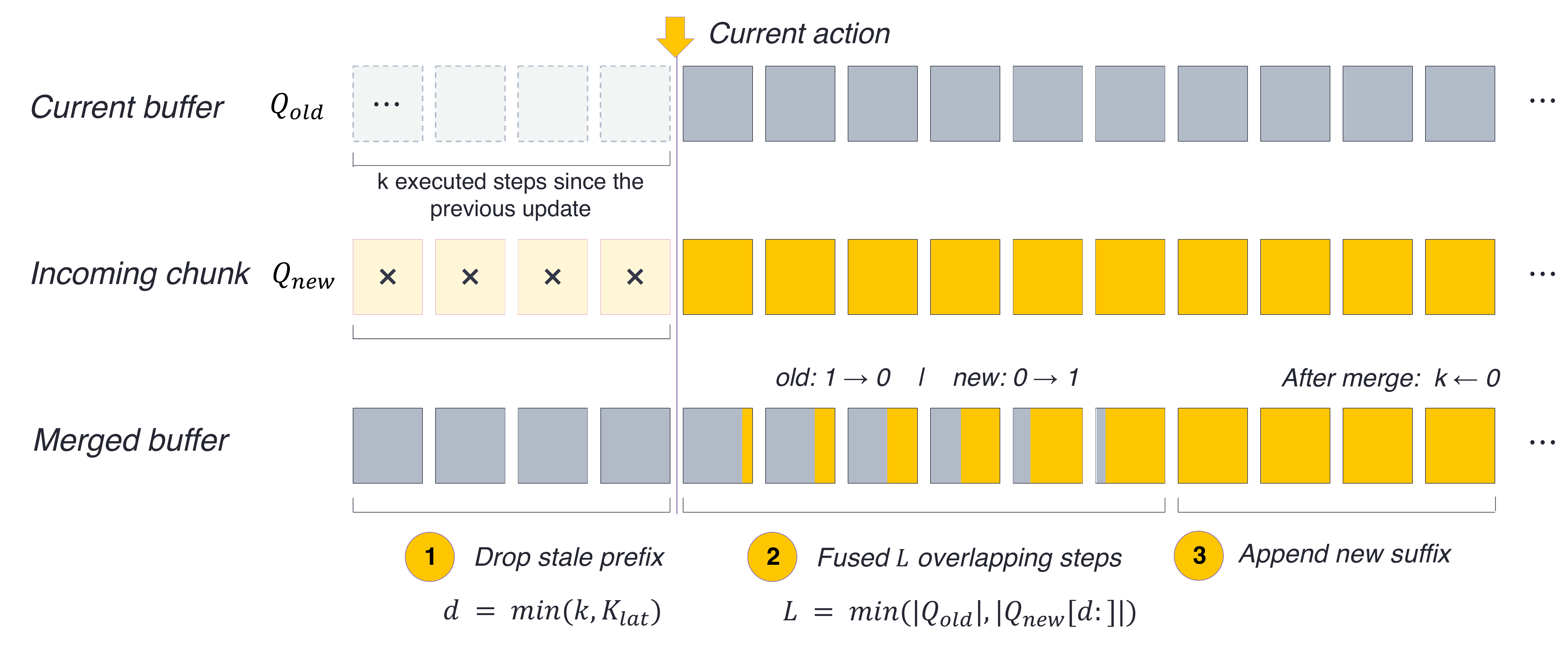}
    \caption{\textbf{Asynchronous action-chunk merging.}
    Predictions made stale during inference are removed from the chunk; the aligned overlap is blended with the residual buffer, and the suffix is appended for continuous execution.}
    \vspace{-5mm}
    \label{fig:async_time}
\end{figure}

\subsection{Asynchronous Policy Execution}
\label{sec:asynchronous_policy_execution}

Action chunks amortize policy inference over multiple control steps, but
inference latency requires coordinating chunk generation with
ongoing execution~\cite{black2025realtime,black2025trainingrtc,zhao2023learning}.
Our EVA-Client based stack~\cite{yang2026evaclient} keeps control continuous by
decoupling the two: a background policy thread predicts a chunk from the latest
observation at target rate $f_\pi$, while the control loop consumes one buffered
command at rate $f_c$.

This concurrency creates a temporal offset. By the time a new chunk
$Q_{\mathrm{new}}$ arrives, the controller has consumed $k$ commands since the
previous successful update. We denote the unexecuted remainder of the current
buffer by $Q_{\mathrm{old}}$. Following the \emph{temporal chunk-wise smoothing}
of $\chi_0$~\cite{sima2026kai0}, we align the two sequences as shown
in~\Cref{fig:async_time}. We first drop
$d=\min(k,K_{\mathrm{lat}})$ leading commands and compute the remaining overlap:

\begin{equation}
    \bar Q_{\mathrm{new}}=Q_{\mathrm{new}}[d{:}],
    \qquad
    L=\min\!\left(|Q_{\mathrm{old}}|,|\bar Q_{\mathrm{new}}|\right).
\end{equation}

Here $K_{\mathrm{lat}}$ is a step-domain cap, set as
$K_{\mathrm{lat}}\approx\lceil f_c\hat\tau_{\mathrm{e2e}}\rceil$, where
$\hat\tau_{\mathrm{e2e}}$ is the delay from observation capture to insertion of
the processed chunk into the buffer, including communication, policy inference,
post-processing, and inverse kinematics.

For each position $i=0,\ldots,L-1$ in the overlapping region, the updated
command is

\begin{equation}
    q_i^{\mathrm{fused}}
    =(1-\alpha_i)q_i^{\mathrm{old}}
      +\alpha_i\bar q_i^{\mathrm{new}},
    \qquad
    \alpha_i=\frac{i}{\max(L-1,1)}.
\end{equation}

The unused new suffix fills the future buffer, and $k$ is reset after a
successful update.

The runtime parameters $f_c$, $f_\pi$, and $K_{\mathrm{lat}}$ jointly govern
buffer consumption and responsiveness. When the inference pipeline meets its
target rate, roughly $f_c/f_\pi$ commands are consumed between updates.
If policy inference and sequential inverse kinematics together take longer
than $1/f_\pi$, the actual update rate falls below $f_\pi$, and more buffered
commands are consumed before the next update. Updates that arrive too slowly
can exhaust the buffer, whereas overly frequent updates repeatedly restart
blending from the current buffer and can slow the physical response.
The cap $K_{\mathrm{lat}}$ balances removal of stale commands against the
remaining executable horizon. The complete procedure and deployment settings
are provided in~\Cref{sec:asynchronous_policy_execution_details}.

\subsection{Event-Driven Embodied Agent}
\label{sec:event-driven-execution}

The chunk merger stabilizes local policy execution but does not preserve information across subtasks or determine when a high-level plan should change. The embodied agent addresses these requirements through explicit task state, persistent event memory, and event-triggered planning.

\textbf{Task decomposition and execution.} Given a user goal $g$, initial scene context $s_0$, and skill library $\mathcal{S}$, the Planner first produces an ordered sequence of executable subtasks $Q^{(0)}=\mathcal{P}_{\mathrm{plan}}(g,s_0,\mathcal{S})$. During execution, the agent maintains the task state $X_t=(q_t,R_t)$, where $q_t$ is the active subtask and $R_t$ is the remaining sequence. The active subtask serves as the language instruction to the low-level policy, which generates actions as $\pi_{\mathrm{VLA}}(o_t,q_t)$. It remains active until the agent observes sufficient evidence of completion, failure, or a change in task feasibility. This state representation lets the VLA policy execute continuously while the Planner operates only at task-level transition points.

\textbf{Visual monitoring.} The Monitor tracks local execution by mapping a short observation window $W_t$ and the active subtask to a structured caption $C_t=\mathcal{M}(W_t,q_t)$. Conditioning on $q_t$ focuses the description on relevant objects, spatial relations, interactions, and execution progress. The temporal window captures changes such as grasping, placement, or object removal that may be ambiguous in a single frame. The Monitor does not decide whether a subtask has finished or modify $X_t$; it only produces visually grounded evidence for event construction.

\textbf{Event construction and memory.} Monitor outputs enter a caption buffer in observation-time order. At each memory update, $B_t$ contains the captions that have not yet been processed, and the persistent history $\mathcal{H}_{<t}$ stores ordered event summaries together with their supporting keyframes. The Memory Manager converts the new captions into an event and determines whether that event warrants high-level reassessment:
\begin{equation}
(E_t,z_t)
=\mathcal{G}(B_t,\mathcal{H}_{<t},X_t,g),
\qquad z_t\in\{0,1\}.
\label{eq:event-trigger}
\end{equation}

When the new evidence continues the same interaction, it is merged into the latest memory entry; a distinct state change creates a new entry. For each event, the keyframe set $K_t$ is obtained by uniformly sampling frames from the observation windows associated with its captions. Memory insertion is independent of planner activation: routine observations can be recorded with $z_t=0$, while subtask completion, execution failure, and changes in task preconditions set $z_t=1$. This separation preserves information that may become relevant to a later instruction without repeatedly invoking the Planner over an unchanged state.

\textbf{Event-triggered planning.} The Planner performs two related operations. $\mathcal{P}_{\mathrm{plan}}$ constructs or revises a subtask sequence from the goal, scene context, task state, and available skills. When $z_t=1$, $\mathcal{P}_{\mathrm{decide}}$ instead evaluates the current event, its keyframes, relevant entries retrieved from $\mathcal{H}_{<t}$, the task state $X_t$, and the user goal $g$. The resulting decision may preserve the active subtask, advance to the next one, wait for a temporarily unsatisfied precondition, or invoke $\mathcal{P}_{\mathrm{plan}}$ to regenerate the remaining sequence. The Planner judges progress from the observed task state rather than requiring the complete action trajectory. When the event or its visual evidence is inconclusive, it leaves $X_t$ unchanged.

\textbf{Asynchronous coordination.} The Monitor, Memory Manager, Planner, and VLA policy operate at different rates. Multiple Monitor requests may run concurrently, so their outputs can arrive out of order; captions are therefore committed according to observation time before event construction. Planner inference does not pause monitoring or low-level execution, and captions produced during planning remain buffered for the next memory update. Task-state mutations are serialized after each Planner decision, preventing concurrent decisions from modifying the same execution state. The system consequently preserves continuous perception and control while keeping event memory and task transitions temporally consistent.

%% file: sections/06-experiments.tex
\section{Experiments}

\input{tables/robotwin}
\subsection{Simulation}

\paragraph{RoboTwin2.0.}
\label{sec:exp:robotwin}

RoboTwin2.0~\citep{robotwin2} is a bimanual desktop-manipulation benchmark comprising 50 tasks with a 14-dimensional joint-space action interface. Each task provides 50 demonstrations in the clean-scene split (\texttt{demo\_clean}) and 500 demonstrations in the randomized split (\texttt{demo\_randomized}), which varies the background, table clutter, table height, and lighting. We jointly fine-tune all 50 tasks end-to-end from the stage-2 checkpoint using native-resolution inputs. Evaluation covers 100 episodes per task on each split, resulting in 5,000 episodes per split. We evaluate the VLA policy alone, without the agent system or RL post-training.

As shown in \Cref{tab:robotwin-main}, Meituan-Robotics-0 achieves the highest overall success rate
among the compared methods at 93.6\%, with 93.2\% on Clean
and 93.9\% on Randomized, a difference of 0.7 percentage points. The comparison with ABot-M0 is particularly informative: both methods predict clean actions and optimize a velocity-space loss, yet MR0 achieves 8.0 points higher overall success. Since MR0 adapts a shared stage-2 checkpoint learned from heterogeneous data and embodiments, these results demonstrate \textbf{\textit{effective transfer to precise bimanual control and show that action-head parameterization alone does not determine downstream performance.}}

\paragraph{RoboDojo.}

RoboDojo~\cite{chen2026robodojo} evaluates generalist manipulation policies on 42 bimanual simulation tasks spanning Generalization, Precision, Long-Horizon, Memory, and Open-vocabulary instruction following. We extend the base MR0 policy with an initial-scene reference and motion history to provide temporal context.
The reference uses the first main-camera image, while a frozen CD-LAM encoder~\cite{wei2026cdlam} extracts latent actions from historical image pairs.
We aggregate these features into one-second intervals over the latest twenty seconds and project them into the backbone with absolute and relative time encodings.
Together with the reference image, they complement the current observations and task instruction while preserving the original action-learning objective.
The embodied-agent system and RL-based post-training are not used.
\Cref{tab:robodojo} reports Score and Success Rate.

With approximately 13.6k hours of embodied training data, the history-augmented MR0 achieves an average Score of 21.68 and an SR of 15.73\%, ranking fifth by average success rate among the compared VLA models.
\textbf{\textit{These results highlight competitive overall performance at a relatively modest data scale, suggesting effective use of heterogeneous pre-training data.}} 
On the Open split, MR0 achieves a Score of 4.28 and an SR of 3.83\%, outperforming Xiaomi-Robotics-1, GalaxeaVLA (G0.5), and DM0.5 on both metrics. Despite this relative advantage, the policy can still fall back on familiar actions when given an unfamiliar instruction. On \texttt{Align\_blocks}, for example, it sometimes grasps the box instead of aligning it.

\input{tables/robodojo}

\subsection{Real-Robot Evaluation}
\label{sec:real_robot_evaluation}

We evaluate Meituan-Robotics-0 at two complementary levels. The policy evaluation measures manipulation performance and assesses the contributions of HIL data and, for Ethernet insertion, replaced policy segments retained by Rollback. The agent evaluation examines two history-dependent capabilities: retrieving information no longer available in the current observation and resuming an interrupted task when its execution conditions are restored.

\subsubsection{Experimental Setup}
\begin{figure*}[!htbp]
    \centering
    \includegraphics[width=0.9\textwidth,pagebox=cropbox]{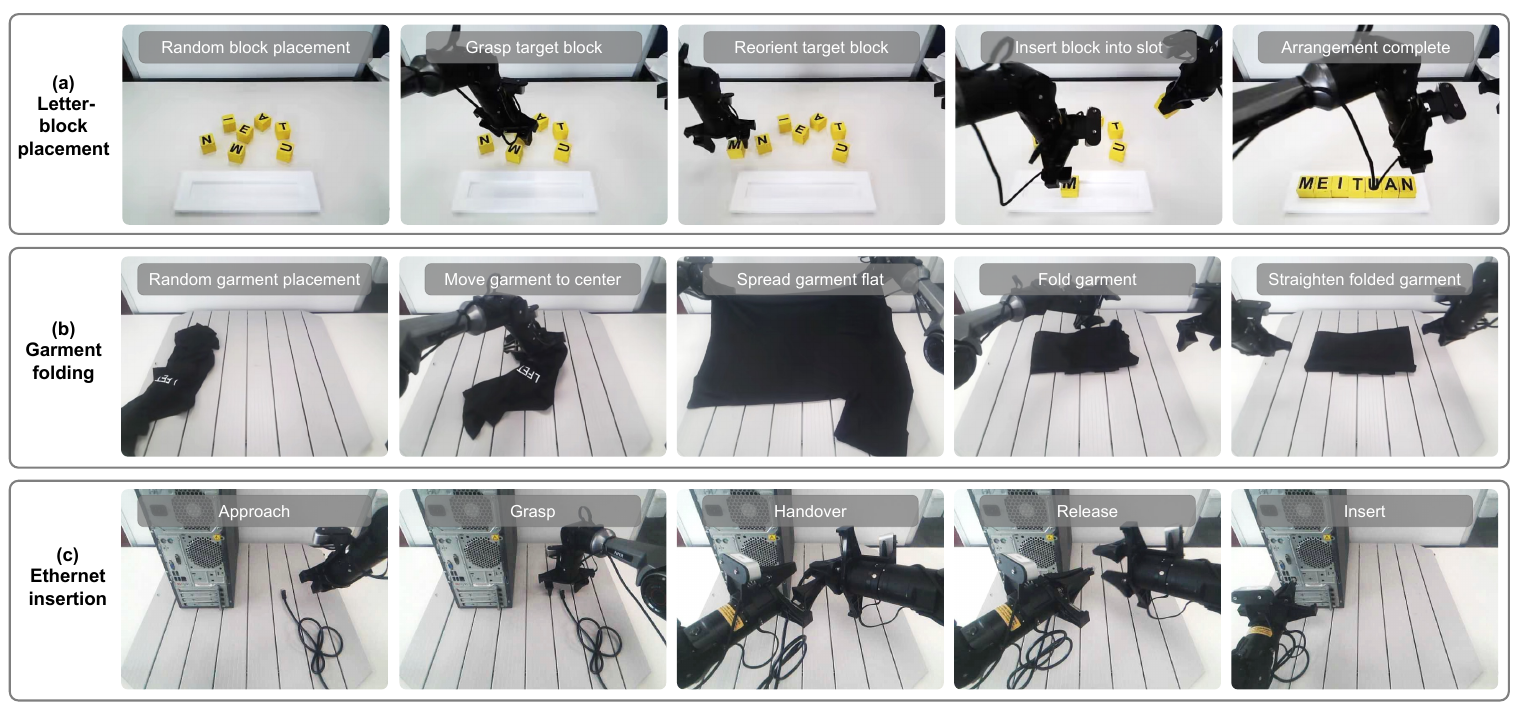}
    \caption{\textbf{Real-robot evaluation tasks.} Representative stages of (a) letter-block placement, (b) garment folding, and (c) Ethernet insertion.}
    \label{fig:real_robot_tasks}
    \vspace{-5pt}
\end{figure*}

\textbf{Tasks and metrics.} We evaluate three tasks covering semantic grounding, deformable-object manipulation, and precise bimanual control (\Cref{fig:real_robot_tasks}). We report success rate (SR) and a task-specific progress score (PS):
\begin{itemize}[
    leftmargin=1.4em,
    itemsep=3pt,
    topsep=3pt,
    parsep=0pt,
    partopsep=0pt
]
    \item \textbf{Letter-block placement.}
    The robot identifies, reorients, and places lettered blocks into target grooves to spell ``MEITUAN''. Success requires all seven letters in the correct order and orientation; $\mathrm{PS}_{\mathrm{block}}=k/7$, where $k$ counts consecutive correct placements before the first error.

    \item \textbf{Garment folding.}
    The robot spreads and folds a garment into a target configuration. Success requires completion within 120 seconds. Evaluation continues for up to 300 seconds. Later completion counts as failure but retains its actual completion time for scoring. Trials unfinished at 300 seconds receive $T=300$. The progress score is $\mathrm{PS}_{\mathrm{fold}}=\operatorname{clip}((300-T)/210,0,1)$, where $T$ is the completion time in seconds.

    \item \textbf{Ethernet insertion.}
    The robot grasps a cable, transfers it between arms, and aligns and inserts the connector into a port. Success requires full insertion; $\mathrm{PS}_{\mathrm{ethernet}}=m/3$, where $m$ counts completed milestones: grasping, handover, and insertion.
\end{itemize}

Each policy is evaluated autonomously for 30 trials per task, with matched initial conditions and the same rollout protocol within each comparison. We average SR and PS over evaluation trials.

\textbf{Post-training data.} All tasks use successful teleoperated demonstrations. Letter-block placement and garment folding add corrections collected through standard HIL. \Cref{tab:real_robot_data} summarizes the data; setup and data preparation are detailed in~\Cref{sec:real_robot_data_preparation}.

\begin{table}[!htbp]
\centering
\small
\caption{\textbf{Real-world post-training data.} Demo counts teleoperated demonstrations; HIL / Rollback counts correction episodes collected with each task's mechanism. Training Samples counts valid 50-step action chunks after preprocessing. The Ethernet row applies to the HIL scaling and Rollback comparisons; the MR0 vs. $\pi_{0.5}$ comparison uses over 1,400 Ethernet episodes.}
\label{tab:real_robot_data}
\begin{tabular}{lrrrr}
\toprule
Task & Demo & HIL / Rollback & Total & Training Samples \\
\midrule
Letter-block placement & 500 & 200 & 700 & $\sim$900K \\
Garment folding        & 400 & 400 & 800 & $\sim$1.1M \\
Ethernet insertion     & 200 & 450 & 650 & $\sim$300K \\
\bottomrule
\end{tabular}
\end{table}

\textbf{Compared policies.} We compare MR0 with $\pi_{0.5}$~\citep{black2025pi05} after SFT and HG-DAgger post-training on all three tasks. Both backbones use the same task data, observation and action representations, preprocessing pipeline, and inference settings. On Ethernet insertion, we compare MR0 trained with HG-DAgger, RECAP, and CFG-DAgger using the same demonstrations and Rollback episodes. All three methods are trained from the same mid-training checkpoint, $\pi_{\mathrm{mid}}$.

\subsubsection{Policy Evaluation}
\label{sec:real_robot_policy_evaluation}

\begin{figure*}[t]
\centering
\includegraphics[width=\textwidth]{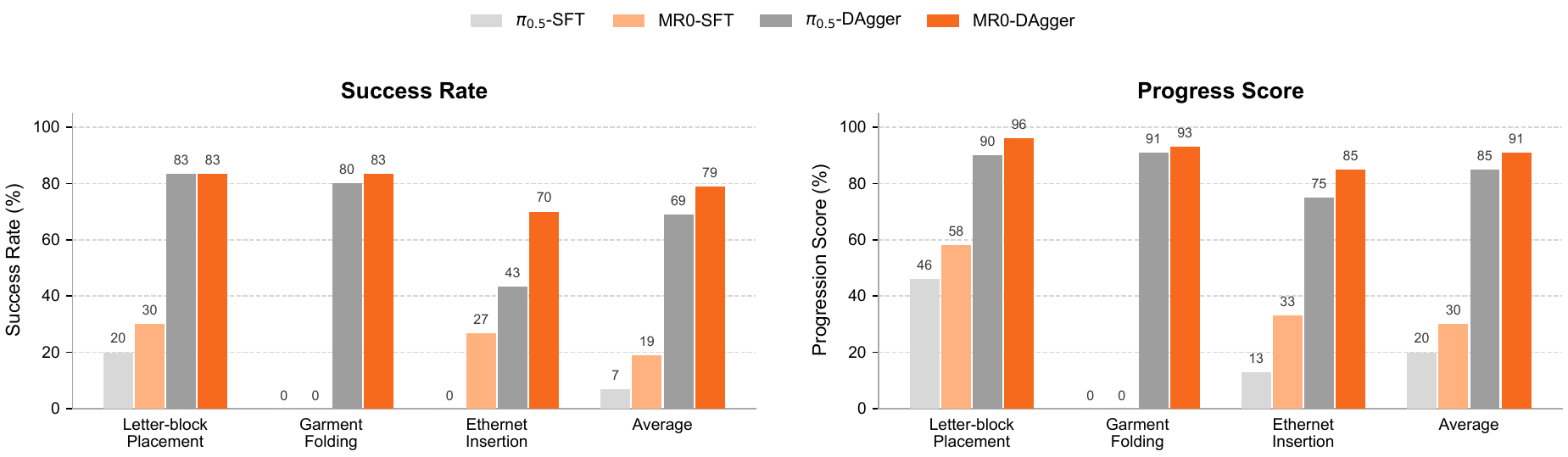}
\caption{Real-robot SR and PS of $\pi_{0.5}$ and MR0 after SFT and HG-DAgger post-training. PS is displayed as a percentage in the figures.}
\label{fig:real_robot_main}
\end{figure*}

\textbf{Main comparison.} Under matched HG-DAgger post-training, MR0 and $\pi_{0.5}$ achieve the same SR on letter-block placement, while MR0 completes more placements before failure (\Cref{fig:real_robot_main}). Garment-folding SR is also similar, with slightly higher PS for MR0. On Ethernet insertion, MR0 achieves the largest improvements in both SR and PS over $\pi_{0.5}$.

\begin{figure*}[!htbp]
\centering
\includegraphics[width=\textwidth]{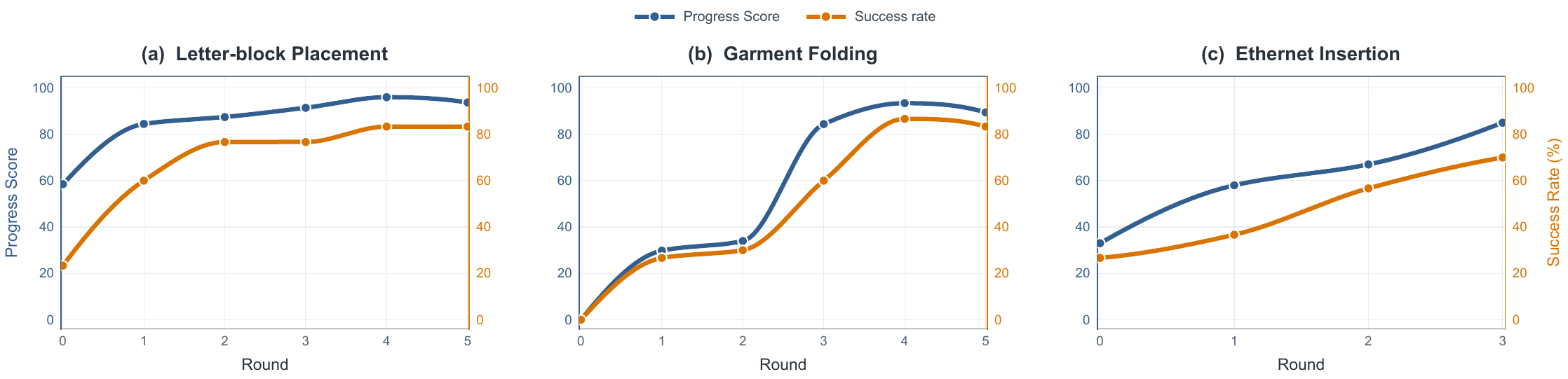}
\caption{Effect of accumulating HIL data on the three tasks. The initial point uses demonstrations only. For each task, each subsequent round adds the same number of HIL episodes. Each policy is trained independently from the same mid-training checkpoint.}
\label{fig:hil_scaling}
\end{figure*}

\textbf{Effect of HIL data.} HIL corrections substantially improve both backbones (\Cref{fig:real_robot_main}). Neither SFT policy succeeds on garment folding, highlighting the limited coverage of failure and recovery states in successful demonstrations alone. As HIL data increase (\Cref{fig:hil_scaling}), letter-block placement and garment folding improve rapidly before plateauing, while Ethernet insertion continues to improve. The gains from additional corrections therefore depend on the task and existing data coverage.


\begin{figure}[t]
    \centering
    \includegraphics[width=0.90\linewidth]{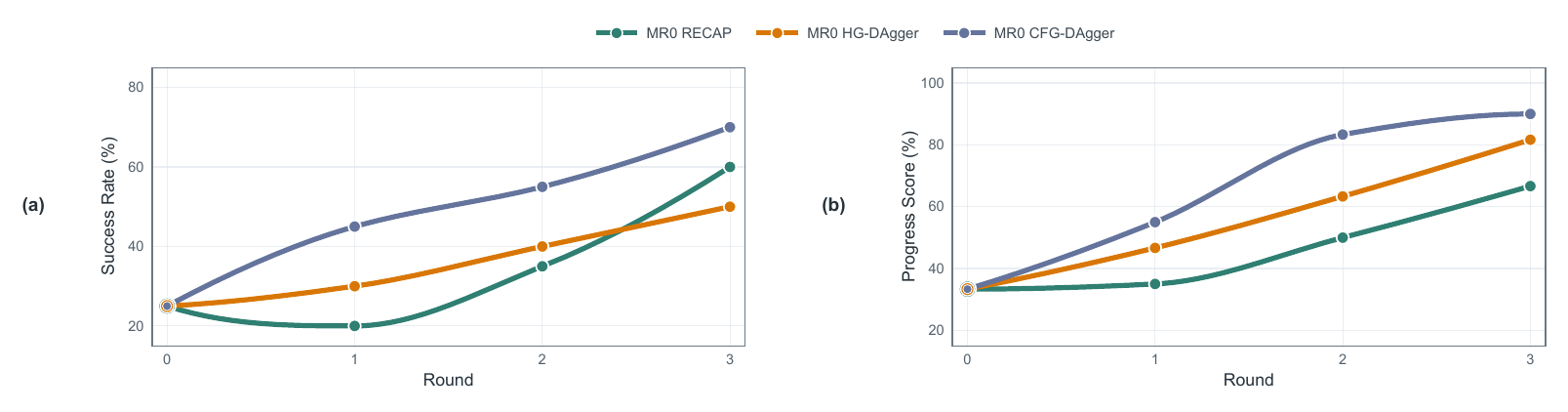}
    \caption{\textbf{Ethernet insertion performance across Rollback collection rounds.} The demonstration set is fixed. Each collection round adds the same number of Rollback episodes. Round~0 is a shared SFT reference.}
    \label{fig:dagger_vs_recap}
\end{figure}

\textbf{Learning from Rollback pairs.} CFG-DAgger achieves higher SR and PS than both baselines after each collection round (\Cref{fig:dagger_vs_recap}). With the full Rollback dataset, CFG-DAgger reaches 0.70 SR and 0.90 PS, compared with 0.60 SR and 0.67 PS for RECAP. CFG-DAgger uses direct segment labels from Rollback and requires neither value learning nor advantage labeling.

\subsubsection{Embodied Agent Evaluation}
\label{sec:embodied_agent_evaluation}

The policy experiments measure execution within a single task, but do not test whether the system can recover information that has disappeared from the current observation or preserve an objective across an interruption. We therefore conduct two controlled qualitative deployments: cross-task event retrieval and condition-aware task resumption.

\textbf{Agent setup.} The Monitor, Memory Manager, and Planner use Qwen3.5-35B-A3B and share an SGLang inference pool with four instances. The Monitor runs at 2~Hz over an eight-frame, two-second observation window. Each persistent event stores a semantic summary and four to eight archived keyframes.

\begin{figure}[!htbp]
\centering
\includegraphics[width=\linewidth]{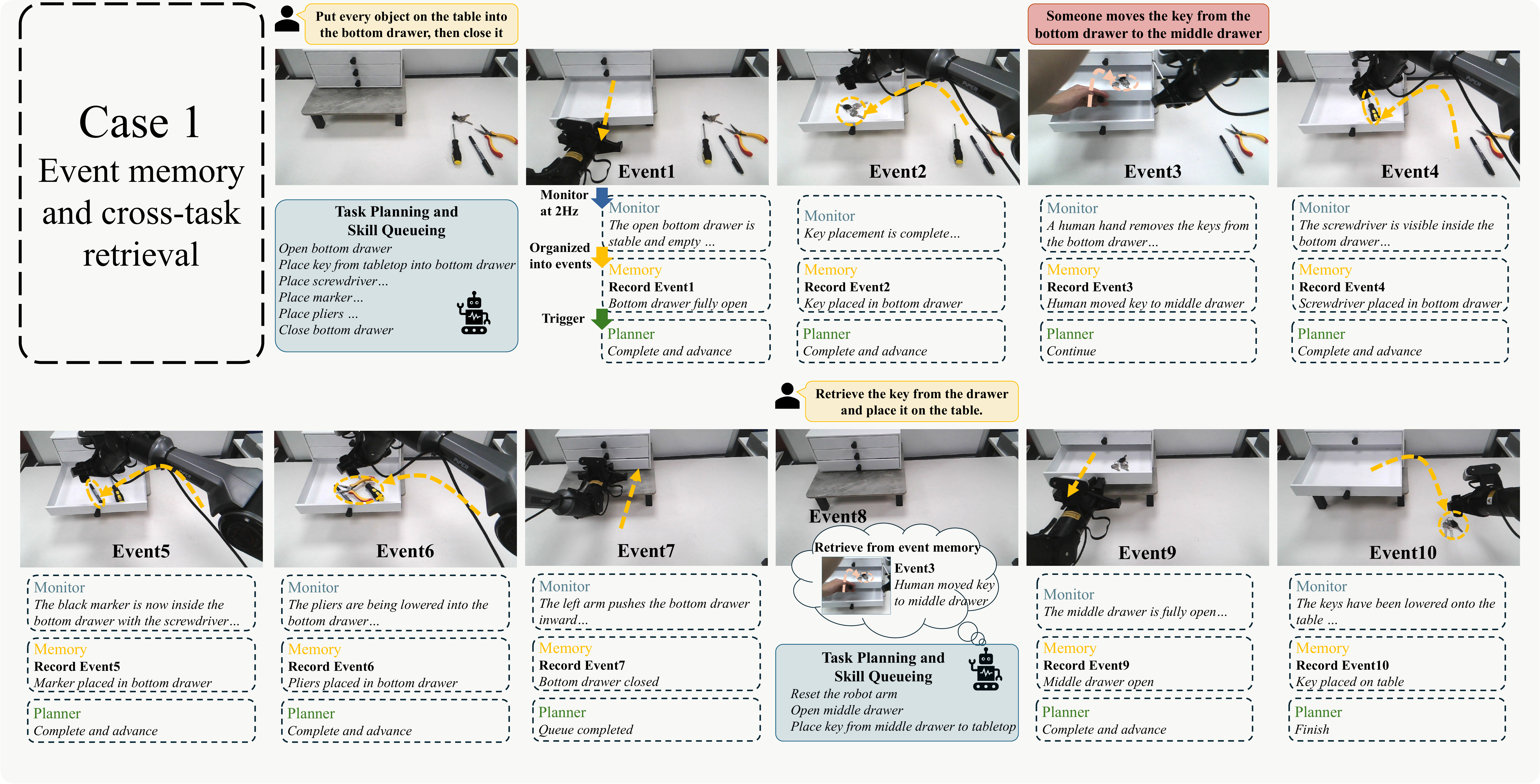}
\caption{\textbf{Event memory and cross-task retrieval.} While screwdriver placement remains active, a person moves the previously placed key from the bottom drawer to the middle drawer. The stored event is later retrieved to locate the key for a new instruction.}
\label{fig:agent_exp1}
\end{figure}

\textbf{Cross-task event retrieval.}
In~\Cref{fig:agent_exp1}, the initial instruction is to place the tabletop objects in the bottom drawer and close it. After the key has been placed and the screwdriver subtask has begun, a person moves the key to the middle drawer. The Monitor captures this state change, and the Memory Manager stores the event with its keyframes. Because the active screwdriver subtask remains feasible, execution continues through the remaining queue and the bottom drawer is closed. This case evaluates event retention and later retrieval, rather than the preservation of all previously satisfied goal conditions under external intervention. When the user later requests the key, neither the closed cabinet nor the empty table reveals its location. The agent retrieves the stored event, identifies the middle drawer as the key's last observed location, and generates a plan to recover it. This case shows that event summaries and archived visual evidence can support a new task when the relevant state is no longer directly observable.

\begin{figure}[!htbp]
\centering
\includegraphics[width=\linewidth]{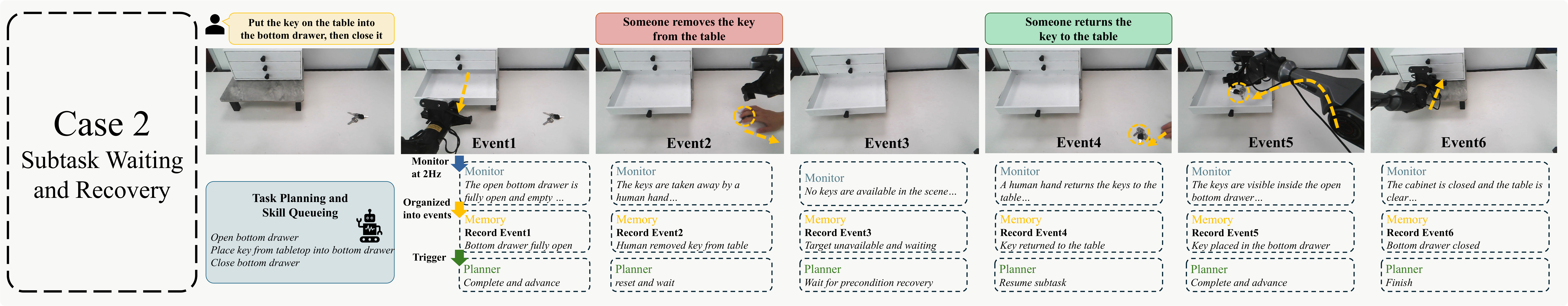}
\caption{\textbf{Subtask waiting and condition recovery.} Removing the key makes the active placement subtask temporarily infeasible. The agent retains the pending objective and resumes execution after the key returns.}
\label{fig:agent_exp2}
\end{figure}

\textbf{Interruption and condition recovery.}
In~\Cref{fig:agent_exp2}, the agent is instructed to place the key in the bottom drawer and close it. A person removes the key after the drawer is opened but before the grasp is completed. The Monitor detects that the target is unavailable, while the Memory Manager records the condition change. The Planner retains the pending placement subtask and enters a waiting state rather than discarding or replanning the objective. When the key returns, a recovery event triggers reassessment against the latest observation, after which the robot places the key and completes the original instruction. This case demonstrates that persistent task state can preserve intent during temporary infeasibility and resume execution when the missing condition is restored.

These qualitative deployments illustrate two uses of persistent context: \textbf{\textit{ retrieving information for a later task and preserving an objective during temporary infeasibility}}. Event memory supports retrieval when relevant state is no longer observable, while persistent task state allows execution to resume after the required conditions are restored.

%% file: tables/robotwin.tex
\begin{wraptable}{r}{0.4\textwidth}
    \vspace{-13pt}
    \centering
    \captionsetup{
        font=small,
        skip=3pt
    }
    \caption{
        \textbf{RoboTwin~2.0 results.}
        Success rate (\%) over 50 tasks and 100 episodes per task in each
        split. Overall is the mean of Clean and Randomized; ``--'' denotes
        an unreported result. Bold and underlining indicate the best and
        second-best results.
    }
    \label{tab:robotwin-main}

    \footnotesize
    \setlength{\tabcolsep}{3.2pt}
    \renewcommand{\arraystretch}{1.02}

    \begin{adjustbox}{max width=\linewidth}
        \begin{tabular}{@{}lccc@{}}
            \toprule
            \rowcolor{groupgray}
            \textbf{Method}
            & \textbf{Clean}
            & \textbf{Random.}
            & \textbf{Overall} \\
            \midrule

            $\pi_0$~\citep{black2024pi0}
            & 65.9 & 58.4 & 62.2 \\

            $\pi_{0.5}$~\citep{black2025pi05}
            & 82.7 & 76.8 & 79.8 \\


            X-VLA~\citep{xvla2026}
            & 72.8 & 72.8 & 72.8 \\

            ABot-M0~\citep{abotm0}
            & 86.1 & 85.1 & 85.6 \\

            Qwen-VLA-Instruct~\citep{qwenvla2026}
            & 86.1 & 87.2 & 86.7 \\

            LingBot-VLA~\citep{lingbotvla2026}
            & 86.5 & 85.3 & 85.9 \\

            Motus~\citep{motus2025}
            & 88.7 & 87.0 & 87.8 \\

            StarVLA~\citep{starvla2026}
            & 88.7 & 87.8 & 88.3 \\

            InternVLA-A1~\citep{internvla_a1_2026}
            & 89.4 & 89.6 & 89.5 \\

            Being-H0.7~\citep{beingh07_2026}
            & 90.2 & 89.6 & 89.9 \\

            Fast-WAM~\citep{fastwam2026}
            & 91.9 & 91.8 & 91.8 \\

            LingBot-VA~\citep{lingbotva2026}
            & 92.9 & 91.5 & 92.2 \\

            InternVLA-A1.5~\citep{internvla_a15_2607}
            & \underline{93.3} & \underline{93.0} & 93.2 \\

            GalaxeaVLA (G0.5)~\citep{liu2026g0}
            & \textbf{93.7} & 92.8 & \underline{93.3} \\

            \textbf{Meituan-Robotics-0}
            & 93.2 & \textbf{93.9} & \textbf{93.6} \\

            \bottomrule
        \end{tabular}
    \end{adjustbox}

    \vspace{-20pt}
\end{wraptable}

%% file: tables/robodojo.tex
\begin{table*}[t]
\centering
\caption{
\textbf{RoboDojo simulation results.}
We compare selected WAM and VLA entries from the official leaderboard.
Within each group, models are ordered by increasing average SR.
Results are reported as Score/SR (\%).
\emph{Embodied Data} reports disclosed embodied trajectory/video training
hours before RoboDojo adaptation, including inherited pre-training where
traceable; -- denotes an unavailable verified duration.
Bold indicates the best Score or SR among the displayed models in each column.
Baseline results follow the official leaderboard; MR0 results are the
equally weighted mean across three evaluation seeds.
Generalization averages the standard and randomized settings.
}
\label{tab:robodojo}
\vspace{1mm}
\small
\setlength{\tabcolsep}{3pt}
\renewcommand{\arraystretch}{1.10}
\begin{adjustbox}{max width=\textwidth}
\begin{tabular}{@{}lccccccc@{}}
\toprule
\textbf{Model} & \textbf{Embodied Data}
& \textbf{Avg. $\uparrow$}
& \textbf{Generalization $\uparrow$}
& \textbf{Precision $\uparrow$}
& \textbf{Long-Horizon $\uparrow$}
& \textbf{Memory $\uparrow$}
& \textbf{Open $\uparrow$} \\
\midrule
\rowcolor{groupgray}
\multicolumn{8}{@{}l}{\textit{\textbf{WAM}}} \\
InternW0-$\Delta$~\cite{miao2026internw0delta}
& $\sim$23.1k h$^\dagger$
& 30.77/23.91 & 30.09/22.78 & 31.98/23.25 & 46.29/29.33 & 34.67/34.00 & 10.84/10.17 \\

Liber-0 Lite~\cite{rd_liber}
& --
& 29.24/24.23 & 25.36/19.11 & 36.60/31.08 & 45.50/32.92 & 35.68/35.44 & 3.06/2.58 \\

Awomo-0.5~\cite{team2026awomo}
& --
& 35.34/29.64 & 30.31/23.84 & 39.65/33.50 & 40.36/26.75 & 42.27/41.22 & \textbf{24.09}/\textbf{22.92} \\

VPP2~\cite{guo2026video}
& --
& 39.26/32.26 & 36.38/29.50 & \textbf{43.03}/\textbf{35.17} & 56.28/37.00 & 56.01/55.56 & 4.61/4.08 \\

\midrule
\rowcolor{groupgray}
\multicolumn{8}{@{}l}{\textit{\textbf{VLA}}} \\

KinRT~\cite{yang2026route}
& --
& 13.02/8.80 & 14.02/8.61 & 15.65/9.92 & 26.40/18.08 & 4.82/3.56 & 4.23/3.83 \\

Hy-Embodied-0.5-VLA~\cite{zhang2026hy}
& $>$10k h
& 13.07/8.80 & 11.78/8.39 & 13.81/8.00 & 25.74/14.92 & 13.37/12.11 & 0.65/0.58 \\

SimpleMemVLA~\cite{yin2026simplememvla}
& --
& 12.58/9.27 & 6.36/3.95 & 7.42/2.92 & 14.58/5.50 & 33.71/33.22 & 0.85/0.75 \\

Xiaomi-Robotics-1~\cite{team2026xiaomi}
& $>100$k h$^\ddagger$
& 20.07/13.93 & 23.54/17.00 & 26.69/18.83 & 38.39/23.67 & 7.81/6.56 & 3.94/3.58 \\

GalaxeaVLA (G0.5)~\cite{liu2026g0}
& --
& 20.23/14.88 & 18.46/12.83 & 28.25/20.42 & 44.12/32.25 & 8.61/7.33 & 1.73/1.58 \\

\textbf{Meituan-Robotics-0}
& $\sim$13.6k h
& 21.68/15.73 & 16.30/10.50 & 21.24/11.17 & 27.07/13.92 & 39.52/39.22 & 4.28/3.83 \\

DM0.5~\cite{rd_dm05}
& --
& 24.90/19.34 & 15.77/10.95 & 24.82/16.75 & 33.70/19.50 & 47.74/47.44 & 2.43/2.08 \\

Rex-M1 Preview~\cite{rd_rexm1}
& $>100$k h$^\ddagger$
& 33.79/27.44 & 34.05/26.05 & 32.82/23.83 & 46.89/33.00 & 48.53/48.22 & 6.67/6.08 \\

Simate-beta~\cite{rd_simate}
& --
& 33.95/27.96 & 35.09/27.95 & 34.35/26.92 & \textbf{57.84}/\textbf{43.42} & 33.33/33.00 & 9.12/8.50 \\

WALL-OSS-1.0-lite~\cite{rd_wallosslite}
& --
& \textbf{42.59}/\textbf{34.97} & \textbf{46.66}/\textbf{37.94} & 38.16/29.08 & 49.94/32.25 & \textbf{57.42}/\textbf{56.00} & 20.76/19.58 \\

\bottomrule
\end{tabular}
\end{adjustbox}

\vspace{2pt}
\begin{minipage}{\textwidth}
\footnotesize
$^\dagger$InternW0-$\Delta$: 23,072.74 processed hours, including
5,633.77 Ego2Robot hours generated from human demonstrations.
$^\ddagger$ Xiaomi reports more than 100k hours of UMI pre-training data and
approximately 10k hours of post-training data. We report only the
pre-training lower bound because a deduplicated cross-stage total
is unavailable. Rex-M1 inherits the Xiaomi checkpoint; its entry
reports the same inherited lower bound, excluding any additional
undisclosed training data.
\end{minipage}
\end{table*}

%% file: sections/07-conclusion.tex
\section{Conclusion}

We presented Meituan-Robotics-0, a vision-language-action foundation model and embodied system for stationary-base desktop manipulation.
MR0 combines heterogeneous data curation with a three-stage training recipe that connects multimodal pre-training, continuous cross-embodiment control, and real-world policy optimization.
Rollback collection provides human corrections and replaced policy segments as direct supervision for CFG-DAgger, enabling policy improvement without value learning or advantage labeling.
At deployment, asynchronous action-chunk merging supports continuous execution, while an event-driven agent maintains persistent context for task-level planning.
Simulation and real-robot experiments demonstrate competitive manipulation performance, and qualitative deployments illustrate cross-task event retrieval and task resumption after temporary interruptions.
These results support a systems approach in which aligned data, corrective learning, and persistent memory address complementary requirements of desktop manipulation.
Future work will focus on agent and harness systems that interact more closely with VLA policies, using execution feedback to guide planning, memory updates, and policy adaptation for reliable long-horizon manipulation.

%% file: sections/08-contributions.tex
\section{Contributions}\label{sec:contributions}
All people are cataloged \textbf{alphabetically by first name}. ($^\dagger$) indicates the project leader and ($^\ddagger$) indicates the sponsors.
\vspace{10pt}

\noindent\textbf{Contributors:}
\begin{multicols}{5}
\noindent
\raggedright
\begin{spacing}{1.2}
Caicheng Wang \\
Chen Zhao \\
Chengpeng Jin \\
Duo Xu \\
Fengjiao Chen \\
Guanglu Wan$^\ddagger$ \\
Heqing Yang \\
Jia Wang \\
Jiacong Zhou \\
Jie Hu$^\dagger$ \\
Jiyuan Ru \\
Lingfeng Zhang \\
Linjiang Huang \\
Manyuan Zhang \\
Ningchu Liu \\
Shijie Zhou \\
Si Liu \\
Shuo Li \\
Tianyi Zhang \\
Xi Zhang \\
Xin Chen \\
Xuezhi Cao \\
Xunliang Cai$^\ddagger$ \\
Yanbin Zeng \\
Yang Yang \\
Yang Yi \\
Yanli Tan \\
Yifu Xu \\
Yimeng Jia \\
Yuchen Tang \\
Yuchen Xie$^\ddagger$ \\
Yuxiang Zheng \\
Zerwei Chng \\
Zilong Huang \\
Zipeng Ye \\
Ziyuan Qin \\
Ziqin Wang \\

\end{spacing}
\end{multicols}



%% file: sections/appendix.tex
\appendix
\label{appendix_part}
\section{Real-Robot Setup and Data Preparation}
\label{sec:real_robot_data_preparation}

\paragraph{Data collection and recording.} Our bimanual AgileX PiPER setup contains two human-operated leader arms and two corresponding task-executing follower arms. The operator controls the follower arms through leader-arm teleoperation. Each trajectory records three camera streams, the observed follower-arm configuration, and the commands sent to the follower arms. These commands are supplied by the leader arms during human control and by the policy during autonomous execution.

The demonstration dataset $\mathcal{D}_{\mathrm{demo}}$, collected by teleoperation, contains only successful executions. The tasks and data quantities used in our experiments are specified in~\Cref{sec:real_robot_evaluation}.

\paragraph{Real-robot observation and action representation.} The policy receives synchronized images from three RGB cameras: one head camera and two wrist cameras. Each camera captures images at a native resolution of $640\times480$. The policy uses 3 camera images at each timestep, without temporal history.

We use Qwen3.5-VL's Smart Resize during both post-training and inference. It preserves the aspect ratio and bounds the visual token count, resizing images to approximately $352\times256$ before visual encoding.

The policy state is derived from the observed follower-arm configuration, and the action target from the command sent to the follower arms. Both are converted to the same 20-dimensional end-effector representation. For each arm, this representation contains a 3D Cartesian position, a 6D rotation, and a normalized gripper closure value, with 0 denoting fully open and 1 fully closed:

\begin{equation}
\mathbf{x}_t
=
\left[
\mathbf{p}^{L},
\mathbf{r}^{L}_{6\mathrm{D}},
g^{L},
\mathbf{p}^{R},
\mathbf{r}^{R}_{6\mathrm{D}},
g^{R}
\right]
\in
\mathbb{R}^{20},
\qquad
\mathbf{x}_t
\in
\{\mathbf{s}_t,\mathbf{a}_t\}.
\end{equation}

The policy predicts an action chunk:
\begin{equation}
\mathbf{a}_{t:t+H-1}\in\mathbb{R}^{H\times20},
\qquad H=50.
\end{equation}

\paragraph{Static-interval trimming and training windows.} Before constructing post-training samples, we remove sustained intervals in which changes in both observed arm configuration and commanded actions remain below task-specific thresholds. This limits the dominance of hold actions while preserving deliberate slow motion and corrections.

After trimming and source selection, trajectories are split at episode and control-source boundaries. Each training window contains $H=50$ consecutive recorded actions from one retained segment, without tail padding or crossing a takeover boundary. Each window contributes one training sample in~\Cref{tab:real_robot_data}. Method-specific data selection is detailed in~\Cref{sec:posttraining_details}.

\paragraph{Post-training implementation details.} Small random crops and rotations augment the head-camera view; color jitter augments all three views. State and action dimensions are normalized using their 1st and 99th percentiles, with the same training-set statistics used to unnormalize actions during deployment.

\section{Asynchronous Policy Execution Details}
\label{sec:asynchronous_policy_execution_details}

This section gives the complete buffer-update procedure in~\Cref{alg:asynchronous_execution} for the asynchronous
controller summarized in~\Cref{sec:asynchronous_policy_execution}. The control
and inference loops share a locked action buffer and run concurrently.

\begin{table}[t]
    \centering
    \small
    \caption{Configuration of the asynchronous executor with temporal chunk-wise smoothing.}
    \label{tab:async_deployment_config}
    \begin{tabularx}{\linewidth}{@{}l c X@{}}
        \toprule
        Quantity or option & Value & Role \\
        \midrule
        $f_c$ & $30\,\mathrm{Hz}$
            & Command-consumption and publication rate of the control loop. \\
        $f_\pi$ & $3\,\mathrm{Hz}$
            & Target background prediction rate. \\
        $H$ & $50$
            & Action horizon returned by the policy server. \\
        $K_{\mathrm{lat}}$ & $4$
            & Maximum number of leading commands removed from an incoming chunk. \\
        $m_{\min}$ & $8$
            & Minimum old-reference length used for short-buffer padding. \\
        Warm-up queries & $2$
            & Synchronous startup queries before the background loop begins. \\
        \bottomrule
    \end{tabularx}
\end{table}

The client retains all $H=50$ returned actions before latency trimming. Two
warm-up queries initialize the inference and IK pipeline. The buffer is then
cleared, and the current robot state serves as the initial reference.

\begin{algorithm}[t]
\caption{Latency-Aware Asynchronous Policy Execution}
\label{alg:asynchronous_execution}
\begin{algorithmic}[1]
\REQUIRE Action horizon $H$, control rate $f_c$, target inference rate $f_\pi$,
latency cap $K_{\mathrm{lat}}$, minimum old-reference length $m_{\min}$
\STATE Shared state: buffer $Q\gets\emptyset$, counter $k\gets0$, and
$q_{\mathrm{last}}\gets q_{\mathrm{current}}$
\STATE Each pop and each complete buffer update are protected by the same lock.
\STATE Run the following control and inference loops concurrently.
\STATE \textbf{Control loop (at $f_c$):}
\WHILE{the system is running}
    \IF{$Q\neq\emptyset$}
        \STATE $q\gets\MRAlgFunc{PopFront}(Q)$; $k\gets k+1$
        \IF{$Q=\emptyset$}
            \STATE $q_{\mathrm{last}}\gets q$
        \ENDIF
        \STATE $\MRAlgFunc{Publish}(q)$
    \ENDIF
\ENDWHILE
\STATE \textbf{Inference loop (at target $f_\pi$):}
\WHILE{the system is running}
    \STATE $Q_{\mathrm{new}}\gets
        \MRAlgFunc{IK}(\MRAlgFunc{Predict}(\MRAlgFunc{Observe}()))$
        \COMMENT{$|Q_{\mathrm{new}}|=H$}
    \IF{$Q_{\mathrm{new}}=\emptyset$}
        \STATE \textbf{continue}
    \ENDIF
    \STATE $d\gets\min(k,K_{\mathrm{lat}})$
    \IF{$d\geq |Q_{\mathrm{new}}|$}
        \STATE \textbf{continue} \COMMENT{$Q$ and $k$ remain unchanged}
    \ENDIF
    \STATE $\bar Q_{\mathrm{new}}\gets Q_{\mathrm{new}}[d{:}]$
    \IF{$Q=\emptyset\land q_{\mathrm{last}}\neq\varnothing$}
        \STATE $Q_{\mathrm{old}}\gets[q_{\mathrm{last}}]^{m_{\min}}$;
        $q_{\mathrm{last}}\gets\varnothing$
    \ELSIF{$Q=\emptyset$}
        \STATE $Q\gets\bar Q_{\mathrm{new}}$; $k\gets0$
        \STATE \textbf{continue}
    \ELSE
        \STATE $Q_{\mathrm{old}}\gets Q$
        \IF{$|Q_{\mathrm{old}}|<m_{\min}$}
            \STATE Extend $Q_{\mathrm{old}}$ to $m_{\min}$ by repeating its
            terminal command
        \ENDIF
    \ENDIF
    \STATE $L\gets\min(|Q_{\mathrm{old}}|,|\bar Q_{\mathrm{new}}|)$
    \FOR{$i=0$ to $L-1$}
        \STATE $\alpha_i\gets i/\max(L-1,1)$
        \STATE $q_i^{\mathrm{fused}}\gets
                    (1-\alpha_i)q_i^{\mathrm{old}}
                    +\alpha_i\bar q_i^{\mathrm{new}}$
                    \COMMENT{Linear transition}
    \ENDFOR
    \STATE $Q\gets
        [q_0^{\mathrm{fused}},\ldots,q_{L-1}^{\mathrm{fused}}]
        \mathbin{\Vert}\bar Q_{\mathrm{new}}[L{:}]$
    \STATE $k\gets0$
\ENDWHILE
\end{algorithmic}
\end{algorithm}

\section{Real-World Post-Training Details}
\label{sec:posttraining_details}

Supervised fine-tuning on successful demonstrations initializes data collection. Each post-training policy is trained anew from $\pi_{\mathrm{mid}}$ on the accumulated data. For the Ethernet Rollback comparison (\Cref{fig:dagger_vs_recap}), In each round of the three rounds, each of the three methods contributes 50 episodes to a shared pool of 150 new Rollback episodes. Segment annotations identify replaced branches and human corrections. All methods use the interface in~\Cref{sec:real_robot_data_preparation}.

\subsection{HG-DAgger Human Intervention}
\label{sec:real_robot_dagger}

Let $\pi_0$ be the SFT policy trained on demonstrations $\mathcal{D}_{\mathrm{demo}}$. During collection round $k$, the robot executes autonomously under $\pi_{k-1}$ ($g_t=0$) until the operator takes control ($g_t=1$). The operator intervenes when execution stalls, errs, or becomes unsafe, and returns control to the policy after recovery.

Only the human-controlled corrections $\tau^+$ from successful episodes are retained (the samples with $g_t=1$), whether collected through standard HIL or Rollback. Across rounds, these corrections accumulate with the demonstrations to form the cumulative supervised dataset for training the next policy.

HG-DAgger changes data collection and sample selection but retains the flow-matching action loss. All retained SFT and HG-DAgger samples receive equal sampling weight.

\subsection{RECAP Policy Optimization}
\label{sec:recap_details}

\paragraph{Trajectory reconstruction and data selection.}
For each Rollback episode, we reconstruct full trajectories from the autonomous prefix preceding the anchor~(\Cref{fig:rollback_collection}): the corrected continuation $\tau^+$ gives a successful trajectory $(\text{prefix}+\tau^+)$, and a replaced branch $\tau^-$, which terminates at the intervention, gives a failed one $(\text{prefix}+\tau^-)$. Value learning uses the demonstrations together with both trajectory types, while policy training retains only the successful trajectories. The procedure is summarized in~\Cref{alg:recap}.

\begin{algorithm}[t]
\caption{RECAP policy optimization with Rollback trajectories}
\label{alg:recap}
\begin{algorithmic}[1]
\REQUIRE $\pi_{\mathrm{mid}}$, value initialization $V_{\mathrm{init}}$, demonstrations $\mathcal{D}_{\mathrm{demo}}$, Rollback batches $1,\ldots,K$
\FOR{$k=1$ to $K$}
    \STATE Reconstruct trajectories from Rollback batch $k$: $\text{prefix}+\tau^+$ (success) and $\text{prefix}+\tau^-$ (failure at intervention)
    \STATE Fit $V_k$ from $V_{\mathrm{init}}$ on the demonstrations and all trajectories reconstructed through batch $k$, using outcome-derived targets
    \STATE Assign $I_t=1$ to $\tau^+$ (human-correction) chunks; label other training windows using~\Cref{eq:recap_condition_state}
    \STATE Train $\pi_k^{\mathrm{RL}}$ from $\pi_{\mathrm{mid}}$ on the successful $(\text{prefix}+\tau^+)$ trajectories with advantage conditioning
\ENDFOR
\ENSURE Final task policy $\pi_K^{\mathrm{RL}}$
\end{algorithmic}
\end{algorithm}

\paragraph{Outcome rewards and value targets.}
Trajectory outcome labels are converted into per-timestep rewards:

\begin{equation}
r_t=
\begin{cases}
0, & t=T \text{ and the trajectory is labeled successful},\\
-T_{\max}, & t=T \text{ and the trajectory is labeled failed},\\
-1, & \text{otherwise},
\end{cases}
\qquad
R_t=\sum_{t'=t}^{T}r_{t'},
\label{eq:recap_reward_return}
\end{equation}

where $T$ is the terminal timestep of the reconstructed trajectory (the intervention point for a replaced branch) and $T_{\max}$ is the maximum episode length. Nonterminal steps incur unit cost, and a failure label adds a terminal penalty of $T_{\max}$. Normalizing by $D=2T_{\max}$ gives the clipped value target

\begin{equation}
y_t
=
\operatorname{clip}
\left(
\frac{R_t}{D},-1,0
\right)
=
\begin{cases}
\operatorname{clip}\left(-\dfrac{T-t}{D},-1,0\right),
& \text{success},\\[4pt]
\operatorname{clip}\left(-\dfrac{T-t+T_{\max}}{D},-1,0\right),
& \text{failure}.
\end{cases}
\label{eq:recap_value_target}
\end{equation}

Targets lie in $[-1,0]$, with values closer to zero indicating shorter successful completions and the terminal penalty distinguishing failures.

\paragraph{Distributional value prediction.}
The value network predicts a categorical distribution over $B=201$ atoms:

\begin{equation}
p_{\phi}(V\mid o_t,\ell)\in\Delta^B,
\qquad
z_i
=
V_{\min}
+
i\frac{V_{\max}-V_{\min}}{B-1},
\quad i=0,\ldots,B-1,
\label{eq:recap_value_support}
\end{equation}

where $V_{\min}=-1$, $V_{\max}=0$, and the atom spacing is $\Delta z=0.005$. We project a scalar target onto its two neighboring atoms using

\begin{equation}
b_t
=
\operatorname{clip}
\left(
\frac{y_t-V_{\min}}{\Delta z},0,B-1
\right),
\qquad
l=\lfloor b_t\rfloor,
\qquad
u=\lceil b_t\rceil.
\label{eq:recap_value_projection_location}
\end{equation}

When $l<u$, the target masses are $q_l=u-b_t$ and $q_u=b_t-l$. When $l=u$, we set $q_l=1$; all other entries are zero. The value loss and continuous prediction are

\begin{equation}
\mathcal{L}_V
=
-\sum_{i=0}^{B-1}q_i
\log p_{\phi}(V=z_i\mid o_t,\ell),
\qquad
V(o_t,\ell)
=
\sum_{i=0}^{B-1}
p_{\phi}(V=z_i\mid o_t,\ell)z_i.
\label{eq:recap_distributional_value}
\end{equation}

\paragraph{Value-model architecture.}
Following RLinf's RECAP implementation~\citep{rlinf_recap}, a SigLIP2-So400m/14 encoder maps each of the three RGB views to 256 tokens projected to 640 dimensions. An 18-layer Gemma~3 270M backbone encodes these visual tokens together with the instruction.

A separate 18-layer value expert with hidden dimension 256 updates a learnable $[\mathrm{CLS}]$ token. It attends to the vision-language prefix through shared attention keys and values, while the prefix cannot attend back to it. A linear head predicts the value distribution:

\begin{equation}
h_t^{\mathrm{CLS}}\in\mathbb{R}^{256}
\longrightarrow
\operatorname{logits}_t
=
W_Vh_t^{\mathrm{CLS}}+b_V
\in\mathbb{R}^{201}.
\label{eq:recap_value_head}
\end{equation}

The vision encoder and Gemma backbone use pretrained weights. The value expert and head are randomly initialized, while the $[\mathrm{CLS}]$ token is initialized to zero. All components are trained jointly.

\paragraph{$N$-step advantage labeling.}
Transition rewards use the same scale as the normalized value targets:

\begin{equation}
\widetilde{r}_t
=
\begin{cases}
y_t-y_{t+1}, & t<T,\\
y_T, & t=T.
\end{cases}
\label{eq:recap_normalized_reward}
\end{equation}

For a training window starting at frame $t$ with a complete $N$-step horizon, the advantage is

\begin{equation}
A_t^{(N)}
=
\sum_{k=0}^{N-1}\widetilde{r}_{t+k}
+
V(o_{t+N},\ell)
-
V(o_t,\ell).
\label{eq:recap_nstep_advantage}
\end{equation}

If $t+N>T$, rewards are accumulated through termination and the bootstrap value beyond the episode is set to zero. Before condition dropout, human-controlled (HIL) chunks are always assigned $I_t=1$. For non-HIL chunks, we assign a conditioning state $I_t\in\{1,\varnothing\}$ using the advantage threshold:

\begin{equation}
I_t
=
\begin{cases}
1, & A_t^{(N)}\geq\epsilon,\\
\varnothing, & A_t^{(N)}<\epsilon.
\end{cases}
\label{eq:recap_condition_state}
\end{equation}

Here, $N=H=50$ and $\epsilon$ is the 70th percentile of advantages over the aggregated dataset. Each non-HIL training window inherits the conditioning state of its first frame.

\paragraph{Advantage-conditioned policy training.}
Before condition dropout, the conditioning state determines the suffix appended to the task instruction:

\begin{equation}
c_t
=
\begin{cases}
\texttt{Advantage: positive}, & I_t=1,\\
\varnothing, & I_t=\varnothing.
\end{cases}
\label{eq:recap_condition}
\end{equation}

For chunks with $I_t=1$, the positive suffix is dropped with probability $p_{\mathrm{drop}}=0.3$; chunks with $I_t=\varnothing$ always use the base instruction. All retained chunks use the flow-matching action loss.

\paragraph{Action-space classifier-free guidance.}
At inference, we denoise a base prediction $\mathbf{a}_{\varnothing}$ ($I=\varnothing$) and a positive-conditioned prediction $\mathbf{a}_{+}$ ($I=1$) from the same initial noise, then combine the completed action chunks once in action space:

\begin{equation}
\mathbf{a}_{\mathrm{CFG}}
=
\mathbf{a}_{\varnothing}
+
\beta
\left(
\mathbf{a}_{+}
-
\mathbf{a}_{\varnothing}
\right),
\qquad \beta=1.75.
\label{eq:recap_cfg}
\end{equation}

The base branch omits only the advantage condition, retaining the observation and task instruction.

\paragraph{Training configuration.}
\Cref{tab:recap_hyperparameters} summarizes the training configuration.

\begin{table}[!htbp]
\centering
\caption{RECAP value learning, advantage labeling, and policy-training hyperparameters.}
\label{tab:recap_hyperparameters}
\begin{tabular}{ll}
\toprule
Parameter & Setting \\
\midrule
Value support $[V_{\min},V_{\max}]$ & $[-1,0]$ \\
Number of value atoms $B$ & 201 \\
Value learning steps / batch size & 6,000 / 256 \\
Value peak learning rate & $5\times10^{-5}$ \\
Value warm-up steps & 500 \\
Value final learning rate & $1\times10^{-6}$ \\
Advantage horizon $N$ / action horizon $H$ & 50 / 50 \\
Advantage threshold $\epsilon$ & Aggregated-data 70th percentile \\
Policy conditions & Positive and base; no explicit negative \\
Condition-dropout rate $p_{\mathrm{drop}}$ & 0.3 \\
Policy training steps / batch size & 30,000 / 128 \\
Policy peak learning rate & $1\times10^{-5}$ \\
Policy warm-up steps & 10,000 \\
Policy minimum learning rate & $5\times10^{-7}$ \\
Inference guidance scale $\beta$ & 1.75 \\
\bottomrule
\end{tabular}
\end{table}

\subsection{CFG-DAgger Conditional Policy Training}
\label{sec:cfg_dagger_details}

\paragraph{Data and labels.}
In~\Cref{fig:rollback_collection}, each pair $(\tau^-,\tau^+)$ shares an anchor but may diverge afterward; no frame-wise alignment is required. If the scene cannot be restored to an anchor, collection falls back to standard HIL and contributes only positive data.

Unlike RECAP, CFG-DAgger uses these local segments directly, without the autonomous prefix: the positive data are the demonstrations and the human corrections $\tau^+$ (label $I=1$), and the negative data are the replaced segments $\tau^-$ (label $I=0$). Labels describe local segments, independent of episode outcomes.

\paragraph{Conditional policy training.}
A shared policy $\pi_\theta(a_{t:t+H-1}\mid o_t,\ell,I)$ learns positive ($I=1$), negative ($I=0$), and unconditional ($I=\varnothing$) branches using the flow-matching loss. Positive samples (demonstrations and $\tau^+$) use condition dropout of $0.3$ to train the unconditional branch while retaining the task instruction. Negative samples ($\tau^-$) always retain their condition. Other autonomous segments, rollback/reset motions, and segments shorter than $H=50$ are excluded. Valid chunks are sampled without class rebalancing, and each round trains from $\pi_{\mathrm{mid}}$ on the cumulative dataset.

\paragraph{Action-space classifier-free guidance.}
At inference, we denoise a positive prediction $\mathbf{a}_{+}$ ($I=1$) and a negative prediction $\mathbf{a}_{-}$ ($I=0$) from the same initial noise, then combine the completed action chunks once in action space:
\begin{equation}
\mathbf{a}_{\mathrm{CFG}}
=
\mathbf{a}_{-}
+
\gamma\left(\mathbf{a}_{+}-\mathbf{a}_{-}\right).
\label{eq:cfg-dagger-guidance}
\end{equation}
We use $\gamma=1.25$, extrapolating from the positive prediction away from the negative prediction. The unconditional branch is used only during training.